\documentclass{article} % For LaTeX2e
\usepackage{iclr2026_conference, times}

\usepackage{amsmath,amsfonts,bm}

\def\eqref#1{equation~\ref{#1}}
\def\1{\bm{1}}

\DeclareMathAlphabet{\mathsfit}{\encodingdefault}{\sfdefault}{m}{sl}
\SetMathAlphabet{\mathsfit}{bold}{\encodingdefault}{\sfdefault}{bx}{n}

\DeclareMathOperator*{\argmax}{arg\,max}

\usepackage{hyperref}
\usepackage{url}
\usepackage{ragged2e}
\usepackage[utf8]{inputenc} % allow utf-8 input
\usepackage[T1]{fontenc}    % use 8-bit T1 fonts
\usepackage{hyperref}       % hyperlinks
\usepackage{url}            % simple URL typesetting
\usepackage{booktabs}       % professional-quality tables
\usepackage{amsfonts}       % blackboard math symbols
\usepackage{nicefrac}       % compact symbols for 1/2, etc.
\usepackage{amsthm}% theorem and definition environments
\newtheorem{definition}{Definition}
\newtheorem{theorem}{Theorem}
\usepackage{microtype}      % microtypography
\usepackage{xcolor}         % colors
\usepackage{microtype}
\usepackage{graphicx}
\usepackage{caption}
\usepackage{subfigure}
\usepackage{multirow}
\usepackage{amssymb}
\usepackage{booktabs}
\usepackage{nicefrac}       % compact symbols for 1/2, etc.
\usepackage{pifont}         % for \xmark and \cmark
\newcommand{\xmark}{\ding{55}} % cross mark
\usepackage{makecell}
\usepackage{tikz}
\usepackage[table]{xcolor}
\usepackage{amsmath}
\usepackage{makecell}
\usepackage{amsfonts}
\usepackage{wrapfig}
\usepackage{float}
\usepackage{bm}
\usepackage{amsmath}
\usepackage{amssymb}
\usepackage{mathtools}
\usepackage[ruled, vlined]{algorithm2e}
\title{\texttt{OD\textsuperscript{3}}: Optimization-free Dataset Distillation for Object Detection}

\author{Salwa K. Al Khatib$^{1}$$^*$, Ahmed Elhagry$^{1}$$^*$, Shitong Shao$^{2, 1}$$^*$, Zhiqiang Shen$^{1, \dagger}$ \\
$^{1}$Mohamed bin Zayed University of Artificial Intelligence\\
$^{2}$Hong Kong University of Science and Technology (Guangzhou)\\
\texttt{\{salwa.khatib, ahmed.elhagry, zhiqiang.shen\}@mbzuai.ac.ae}, \\ \texttt{sshao213@connect.hkust-gz.edu.cn} \\
}

\iclrfinalcopy % Uncomment for camera-ready version, but NOT for submission.
\begin{document}

\maketitle

\renewcommand{\thefootnote}{\fnsymbol{footnote}}
\footnotetext[1]{Equal contribution.}
\footnotetext[2]{Corresponding author.}

\vspace{-0.15in}
\begin{abstract}
\vspace{-0.15in}

Training large neural networks on large-scale datasets requires substantial computational resources, particularly for dense prediction tasks such as object detection. Although dataset distillation (DD) has been proposed to alleviate these demands by synthesizing compact datasets from larger ones, most existing work focuses solely on image classification, leaving the more complex detection setting largely unexplored. In this paper, we introduce {\bf \texttt{OD\textsuperscript{3}}}, a novel optimization-free data distillation framework specifically designed for object detection. Our approach involves two stages: first, a candidate selection process in which object instances are iteratively placed in synthesized images based on their suitable locations, and second, a candidate screening process using a pre-trained observer model to remove low-confidence objects. We perform our data synthesis framework on MS COCO and PASCAL VOC, two popular detection datasets, with compression ratios ranging from $0.25\%$ to $5\%$. Compared to the prior solely existing dataset distillation method on detection and conventional core set selection methods, {\bf \texttt{OD\textsuperscript{3}}} delivers superior accuracy, establishes new state-of-the-art results, surpassing prior best method by more than 14\% on COCO mAP$_{50}$ at a compression ratio of $1.0\%$. Code is available at \url{https://github.com/VILA-Lab/OD3}.
\end{abstract}

\begin{wrapfigure}{r}{0.5\textwidth}
    \centering
    \vspace{-0.2in}
    \includegraphics[width=0.9\linewidth, height=0.2\textheight]{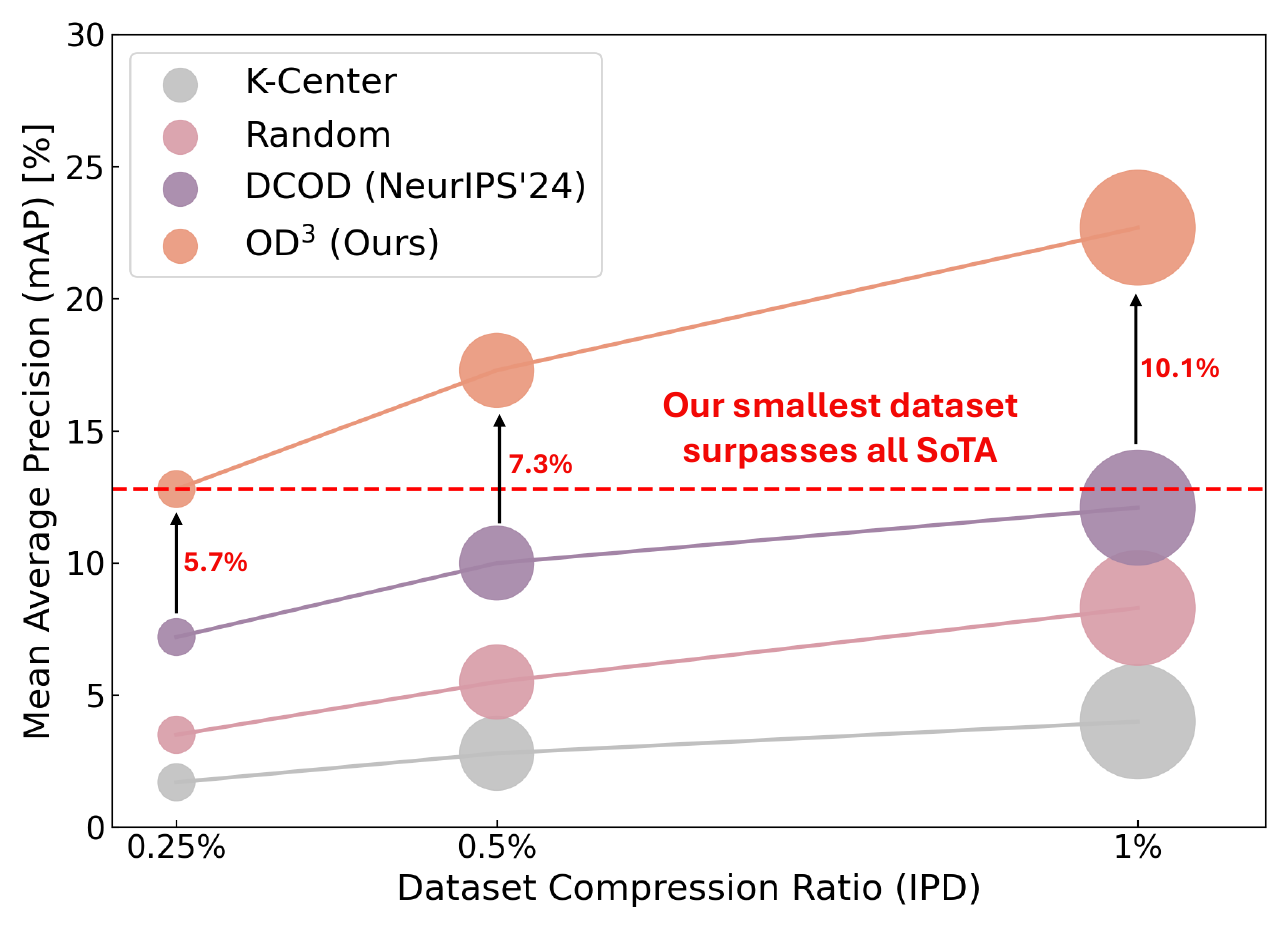}
    \vspace{-0.12in}
    \caption{\textbf{Performance Comparison of {\bf \texttt{OD\textsuperscript{3}}} on COCO.} We compare the mAP performance of our method to others with different compression rates. The upper bound is 39.8\% on the full dataset. }
    \label{fig:ipd_map}
    \vspace{-0.2in}
\end{wrapfigure}

\section{Introduction}
\vspace{-0.15in}
Deep neural networks have achieved remarkable performance across a wide range of computer vision tasks~\citep{he2016deep, ren2015faster, dosovitskiy2020image, kirillov2023segment}, but training these models generally requires substantial computational and data resources. Conventional strategies often involve collecting increasingly large datasets~\citep{imagenet} and training ever larger networks~\citep{dehghani2023scaling} to capture data complexity. This paradigm is particularly evident in object detection~\citep{shao2019objects365}, where the need for rich annotations, such as bounding boxes or even instance masks, can greatly increase dataset sizes and labeling overhead. As a result, there is a growing interest in techniques that enable the creation of smaller, more manageable datasets capable of approximating the performance achieved by training on the original data. One promising direction in this area is dataset distillation (DD), which aims to synthesize {\em condensed} datasets that are significantly smaller yet still effective for training.

\vspace{-0.05in}
The majority of DD approaches have focused on image classification, where each image contains an object or a dominant label. This narrow scope overlooks the complexity and diversity of more demanding tasks, specifically object detection. In contrast to classification, detection requires localizing and identifying multiple instances of potentially different classes in a single image. This jump in task complexity involves learning a mapping from image to label and predicting boxes and class labels for multiple regions within the same image. Consequently, methods that successfully distill datasets for classification often struggle to adapt to the richer problem space of detection.

Another critical distinction lies in the type of supervision and evaluation metrics used in object detection tasks. While classification tasks use labels that can be applied at the image level, detection tasks rely on spatial annotations that align individual objects to bounding boxes, complete with class labels. This requirement introduces additional challenges when creating distilled datasets, as both the geometry (location) and identity (class) of objects must be preserved or effectively synthesized. Approaches that merely compress high-level category information may fail to capture the crucial spatial relationships and visual diversity that define detection tasks.

\begin{figure*}[t!]
    \centering
    \includegraphics[width=\linewidth]{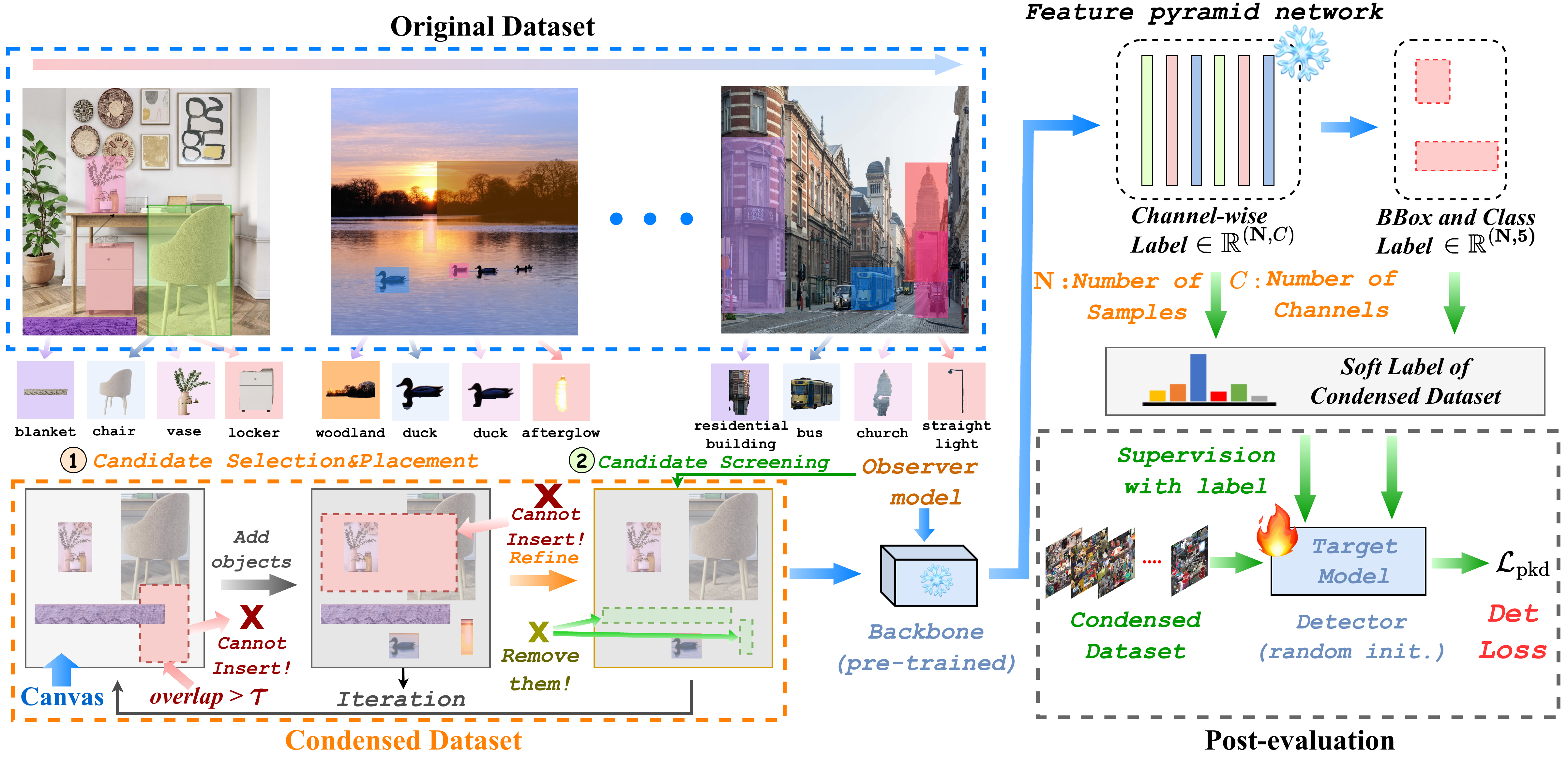}
    \vspace{-0.2in}
    \caption{\textbf{Illustration of the \texttt{OD\textsuperscript{3}} framework.} In initial stage \ding{172}, each object in $\mathbf x_i \in \mathcal{T}$ is assigned a random location in the synthesized image $\mathbf{\hat x}_j \in \mathcal{S}$ for $j=1, ..., \texttt{IPD}$, and its overlap with existing candidates is checked to decide placement.
    After $\texttt{IPD}$ synthesized images are initially constructed, a pre-trained observer model produces predictions for screening. 
    The observer iteratively evaluates the current canvas to identify and remove objects that do not meet expectations to align with the post-evaluation process.
    For final reconstruction, the objects are inserted using their bounding boxes into $\mathbf{\hat x}_j \in \mathcal{S}$. Post-evaluation of $\mathcal{S}$ is carried out by fast distilling knowledge~\citep{shen2022fast} from the observer model to a target network using PKD \citep{pkd} loss on the respective feature pyramid networks.} 
    \label{fig:framework}
    \vspace{-0.15in}
\end{figure*}

\vspace{-0.05in}
In light of these complexities, we propose \textbf{O}ptimization-free \textbf{D}ataset \textbf{D}istillation for Object \textbf{D}etection (\textbf{\texttt{OD\textsuperscript{3}}}), a novel framework explicitly tailored to address the unique challenges of synthesizing small, high-fidelity datasets for object detection. The framework leverages instance-level labels with  scale-aware dynamic context extension (SA-DCE) to reconstruct diverse training images guided by an observer model, which is grounded in two core ideas: (1) an iterative \textit{candidate selection} process that strategically places object instances in synthesized images, and (2) a \textit{candidate screening} process powered by a pre-trained observer model, which discards low-confidence objects. By removing the need for complex optimization procedures in constructing these synthetic images, {\bf \texttt{OD\textsuperscript{3}}} provides a more streamlined and adaptable approach to DD for dense prediction tasks. The main contributions of this work are as follows:
\vspace{-0.1in}
\begin{itemize}
    \item We propose a novel DD framework specifically designed for object detection, named \textbf{\texttt{OD\textsuperscript{3}}}. It involves a two-stage process: \textit{candidate selection}, where masked objects are localized and selected based on minimal overlap, and \textit{candidate screening}, where a pre-trained observer filters unreliable candidates. 
    \vspace{-0.05in}
    \item \textbf{\texttt{OD\textsuperscript{3}}} bridges a crucial gap by extending the concept of dataset distillation beyond the relatively well-explored territory of image classification to the more challenging domain of object detection in a \textit{training-free scheme}. Through a carefully designed process that handles both the spatial and semantic requirements of detection, our framework enables significant reductions in dataset size without sacrificing performance significantly.
    \vspace{-0.05in}
    \item We evaluate our framework on MS COCO with compression ratios ranging 0.25\% to 5\% and on PASCAL VOC from 0.5\% to 2.0\%. The results demonstrate that our framework effectively reduces dataset size while maintaining model accuracy, providing an efficient solution for training object detectors.
    \vspace{-0.05in}
    \end{itemize}

\vspace{-0.05in}
\section{Related Work}
\vspace{-0.05in}
Coreset selection has emerged as one solution for reducing dataset size, primarily in image classification \citep{guo2022deepcore, braverman2019coresets, huang2019coresets}. It shows challenges in object detection, where multiple objects may appear in a single image. Recently, CSOD~\citep{lee2024coreset} introduces Coreset Selection for Object Detection, which selects image-wise and class-wise representative features for multiple objects of the same class using submodular optimization. Similarly, Training-Free Dataset Pruning \citep{trainingfree} addresses dataset pruning for instance segmentation, tackling pixel-level annotations and class imbalances without training. However, these methods often achieve low compression ratios, typically above 20\%. In contrast, our proposed distillation method compresses the original dataset to 0.5\% or less.

Currently, efforts in dataset distillation for object detection remain limited, unlike in image classification \citep{wang2018dataset, cazenavette2022dataset, rded, mtt}. The first framework DCOD \citep{fetchandforge} was proposed for this purpose. DCOD employs a two-stage process: {\em Fetch} and {\em Forge}. During the {\em Fetch} stage, an object detection model is trained on the original dataset to extract essential features for localization and recognition tasks, similar to the squeezing process in SRe$^2$L \citep{yin2024squeeze}. In the {\em Forge} stage, synthetic images are generated via model inversion, embedding required information into the images through uni-level optimization.

\vspace{-0.05in}
\section{Method}
\label{sec:method}
\vspace{-0.05in}
\noindent{\bf {\em Preliminaries}: Dataset Distillation for Object Detection.} The goal of \textbf{\texttt{OD\textsuperscript{3}}} is to compress a large object detection dataset $\mathcal{T}=\{(\mathbf x_i, \{<\!\bm b_{i1}, \bm c_{i1}, \dots\!>\})\}$ $(i=1, \dots, |\mathcal{T}|)$  into a much smaller synthesized dataset $\mathcal{S}=\{(\mathbf{\hat x}_j, \{<\!\bm{\hat b}_{j1}, \bm{\hat c}_{j1}, \dots\!>\})\}$ $(j=1, \dots, |\texttt{IPD}|)$ that maintains the significant characteristics of $\mathcal{T}$ in terms of overall performance, where $\bm b=\{\mathbf x_c, \mathbf y_c, \bm w, \bm h\}$ represents the center of the bounding box and the width and height of the image. Here, $|\mathcal{S}|\ll |\mathcal{T}|$ and \texttt{IPD} is the notion of images per dataset which reflects the compression ratio\footnote{%
\parbox[t]{0.43\textwidth}{%
\justifying
We define \texttt{IPD} (images per dataset) instead of the conventional
\texttt{IPC} (images per class) used in classification tasks, as in object
detection each image can contain multiple objects from different classes.
}}. The performance of a model with weights $\theta_{\mathcal{S}}$ trained on $\mathcal{S}$ should be similar to that of a model with weights $\theta_{\mathcal{T}}$ trained on $\mathcal{T}$, within a small margin $\bm \epsilon_{DD}$. This can be expressed as:

\vspace{-0.15in}

\begin{equation}
sup\{|\mathcal{L_{\theta_\mathcal{T}}} - \mathcal{L_{\theta_\mathcal{S}}}|\}_{(\mathbf{x}_v, \mathbf{y}_v)\in \mathcal{T'}} \leq \bm \epsilon_{DD}
\end{equation}
with $\mathcal{L}$ representing the loss function, and ($\mathbf{x}_v, \mathbf{y}_v$)$\in \mathcal{T}'$ is some test or val set associated with $\mathcal{T}$.

\begin{definition}[Optimization-free dataset distillation for object detection]
\em
Our objective is to collect as much \emph{effective} information as possible on a ``blank canvas'', interpreted as an initially empty image. The information is considered ``effective'' if it contains sufficient high-quality (high-confidence, well-sized) objects.
\end{definition}

\begin{wrapfigure}{r}{0.5\textwidth}
    \centering
    \vspace{-0.42in}
    \includegraphics[width=\linewidth]{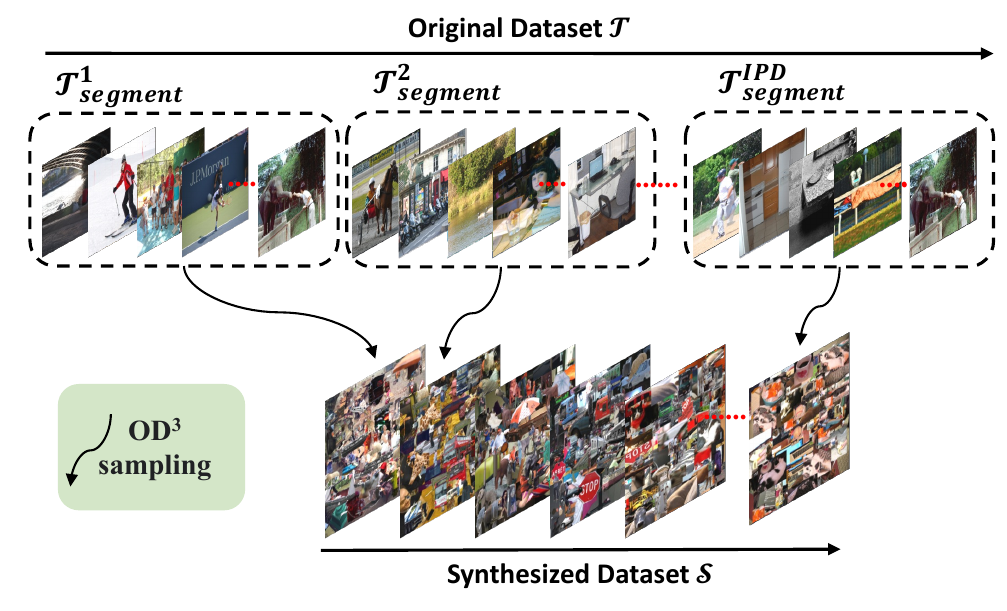}
    \vspace{-0.3in}
    \caption{\textbf{Illustration of our sampling controller.} It ensures that the same object is not placed in different distilled images. The original dataset is divided into \texttt{IPD} segments. Each segment is distilled into a single image, resulting in a compact dataset $\mathcal{S}$.}
    \label{fig:ipd-sampling}
    \vspace{-0.84in}
\end{wrapfigure}
\noindent\textbf{Information Density.}
To quantify how thoroughly a canvas is occupied by valuable objects, we define an \emph{Information Density} function $\Phi(\mathbf x)$:
\begin{equation}
\Phi(\mathbf x) = \frac{g\left(f_\theta(\mathbf x)\right)}{a(\mathbf x)}, 
\end{equation}
where $\mathbf x$ is the current canvas (image) under consideration. $f_\theta(\cdot)$ is a well-trained object detector parameterized by $\theta$. $g(\cdot)$ is a function that aggregates detection confidence scores across all detected objects. $a(\mathbf x)$ denotes the combined area of all detected objects on $\mathbf x$.
Concretely, we instantiate $g(\cdot)$ and $a(\cdot)$ as follows:
% \vspace{-0.1in}
\begin{equation} \label{ID}
\Phi(\mathbf{x})=\frac{\sum_{r=0}^K a\bigl(\bm o_{r}\bigr)q\bigl(\bm o_{r}\bigr)}{\sum_{r=0}^K a\bigl(\bm o_{r}\bigr)}, 
\end{equation}
where $K$ is the total number of objects placed on the canvas $\mathbf x$, $\bm o_{r}$ is the $r$-th object, $a\bigl(\bm o_{r}\bigr)$ represents the area of object $\bm o_{r}$, $q\bigl(\bm o_{r}\bigr)$ is the detector-derived confidence score for $\bm o_{r}$\footnote{In our paper, $i$, $j$ and $r$ are the image index of original dataset, index of distilled image, and object index, respectively.}. Thus, $\Phi(\mathbf x)$ measures how confidently and extensively the canvas is occupied by objects.

\noindent{\textbf{Information Diversity.}}
In addition to confidence and size, we also encourage diversity of objects on the canvas. We define a simple \emph{Information Diversity} $\mathcal{N}(\mathbf x)$ by:
\begin{equation} \label{dviersity}
\mathcal{N}(\mathbf x) = N.
\end{equation}
where $N$ is the number of distinct objects on the canvas $\mathbf x$. Even when a few objects exhibit high confidence, having more \emph{distinct} objects can yield richer training signals, making the distilled data more robust.

% \subsection{\textbf{\texttt{OD\textsuperscript{3}}} Framework}

\subsection{\texorpdfstring{\textbf{\texttt{OD\textsuperscript{3}}} Framework}{OD3 Framework}}
\label{subsec:framework}

\noindent{\bf Overview.}
Unlike prior dataset distillation methods, our approach begins with a {\em blank canvas} as the starting point for generating each new synthetic data sample.
As shown in Fig.~\ref{fig:framework}, the data distillation process first proceeds with a candidate selection stage (orange box, bottom-left), where object instances are extracted from an existing large-scale dataset $\mathcal{T}$. For each image $\mathbf x_i \in \mathcal{T}$, bounding boxes $\left\{\bm b_{i1}, \bm b_{i2}, \ldots, \bm b_{iK}\right\}$ ($K$ is the number of bounding boxes) capture potentially useful object patches. These patches are fed into a localization operation, a random yet controlled placement mechanism that carries out $\bm M$ attempts of inserting each candidate onto a reconstructed canvas without exceeding the overlap threshold. Fig.~\ref{fig:ipd-sampling} shows the sampling strategy that ensures that $|\mathcal{S}| = \texttt{IPD}$ and that objects in $\mathcal{S}$ are all unique. This segment-based synthesis strategy also helps preserve the inter-class balance and intra-class diversity for the distilled data. It yields a large \textit{pool of candidate patches} $\left(\bm b_i, \bm l\right)$ on the canvas, where $\bm l$ is the bounding box label or class. Our illustration also highlights how some patches that fail overlap constraints are discarded.

\noindent{\bf Candidate Screening via Iterative Transfer and Filtering Process.} Once a preliminary reconstructed image is assembled, the process proceeds to the candidate screening / filtering stage. Here, an observer model (a pre-trained detector) performs inference on the partially reconstructed canvas. Its predictions are matched with ground-truth boxes that originated from the bounding boxes inserted into the image. Objects that fail to meet confidence or consistency criteria are removed, refining the canvas into a high-quality, diversified arrangement of objects. As a result, the final reconstructed image $\mathbf{\hat x}_j \in \mathcal{S}$ now contains only those patches that pass the screening process. Also, the bounding box and class annotations associated with these patches are transformed into soft labels, enabling more nuanced supervision in subsequent stages.

\noindent{\bf Soft Label Generation.} 
Logit-based soft labels play a critical role in improving the performance of validation models trained on condensed datasets in image classification~\citep{yin2024squeeze} through KD framework~\citep{hinton2015distilling}. However, when applied to object detection tasks, logit-based soft labels fail to deliver competitive accuracy. This raises the necessity of developing a specialized soft label design tailored explicitly for dataset distillation in object detection. The most typical kind of soft label used in object detection is the output of the feature pyramid network (FPN). This output $\mathbf{y}^\textrm{feat}$ can be defined as $\mathbb{R}^{C \times H \times W}$, where $C$, $H$ and $W$ represent the number of channels, the height of the canvas and the width of canvas, respectively. Once the (feature-based) soft label $\{\mathbf{y}^\textrm{feat}_i\}$ has been obtained, it is employed during the post-evaluation phase and supervised using the following loss function~\citep{shu2021channel}:
\begin{equation}
    \begin{split}
        & \mathcal{L}_\textrm{mse} = \mathbb{E}_{(\mathbf{x}_i, \mathbf{y}^\textrm{feat}_i)} \Vert \mathbf{y}^\textrm{feat}_i - f^\textrm{fpn}(f^\textrm{backbone}(\mathbf{x}_i)) \Vert_2^2, \\
    \end{split}
    \label{eq:mse_loss_obj_detection}
\end{equation}
where $\{(\mathbf{x}_i, \mathbf{y}^\textrm{feat}_i)\}$, $f^\textrm{fpn}$ and $f^\textrm{backbone}$ refer to the condensed dataset, the FPN in the model and the backbone of the model, respectively. However, we observe that this form of soft label is hard to provide sufficient information for detection.

Thus, we consider a channel-wise soft label for enhancing the performance of the evaluated detector. We leverage the simple pearson knowledge distillation (PKD)~\citep{pkd} as a basis for designing the soft label generation mechanism on object detection. Given this, we can give the form of soft label as $\{\frac{f^\textrm{fpn}(f^\textrm{backbone}(\mathbf{x}_i)) - \textrm{mean}(f^\textrm{fpn}(f^\textrm{backbone}(\mathbf{x}_i)))}{\textrm{std}(f^\textrm{fpn}(f^\textrm{backbone}(\mathbf{x}_i)))+\epsilon}\}$, where $\textrm{mean}(\cdot)$, $\textrm{std}(\cdot)$ and $\epsilon$ denote the mean operator in the height and width dimensions, the standard deviation operator in the height and width dimensions, and very small amounts, respectively. Finally, the condensed dataset and its associated soft labels are used to train a new detector initialized randomly to test how well this small synthetic dataset supports the downstream detection task. As shown in the post-evaluation stage, the condensed dataset supervises the target detector training, and PKD used in post-evaluation can be formulated as:

\vspace{-0.1in}
\begin{equation}
        \mathcal{L}_\textrm{mse} = \mathbb{E}_{(\mathbf{x}_i, \mathbf{y}^\textrm{feat}_i)} \Big\Vert \mathbf{y}^\textrm{feat}_i -
        \frac{f^\textrm{fpn}(f^\textrm{backbone}(\mathbf{x}_i)) - \textrm{mean}(f^\textrm{fpn}(f^\textrm{backbone}(\mathbf{x}_i)))}{\textrm{std}(f^\textrm{fpn}(f^\textrm{backbone}(\mathbf{x}_i)))+\epsilon} \Big\Vert_2^2, \\
    \label{eq:pkd_loss_obj_detection}
\end{equation}

\noindent{\bf Scale-aware Dynamic Context Extension.}
We also propose a simple {\em scale-aware dynamic context extension} (SA-DCE) for varying sizes of objects in detection-based dataset distillation as a crucial enhancement that directly addresses the challenges posed by small objects with limited contextual information. Unlike the optimization-based method~\citep{fetchandforge}, which struggles to preserve or amplify contextual cues due to their reliance on fixed gradients and pixel-specific updates, context extension involves intentionally expanding the bounding region around objects. It can be formulated as a function of the object's size:
\begin{equation} \label{extension}
\bm \ell_\text{extension} = F(\bm {\bm o_{i_r}}, \overline{\bm r})  =  \left( 1 - \frac{a({\bm o_{i_r}}) - \overline a_\text{min}}{\overline a_\text{max} - \overline a_\text{min}} \right) \times \overline{\bm r}, 
\end{equation}
\begin{wrapfigure}{r}{0.5\textwidth}
\vspace{-0.2in}
\centering 
\includegraphics[width=\linewidth]{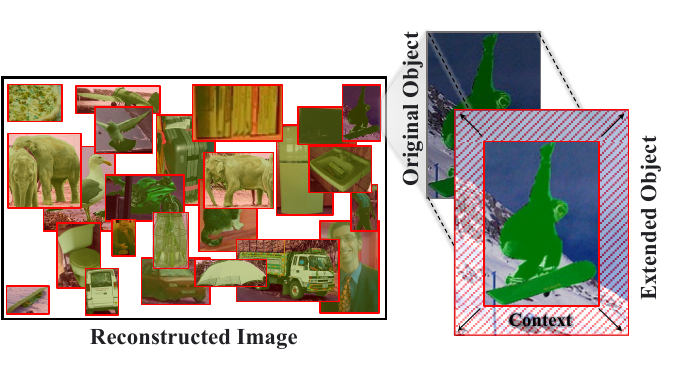} 
\vspace{-0.25in}
\caption{\textbf{Extended bounding box for more object context.} The figure shows a reconstructed image along with an extended object using the SA-DCE function to better capture the object’s context.} 
\label{fig:extended-bbox}
\vspace{-0.25in}
\end{wrapfigure}
where $F$ is the SA-DCE function, $\overline{\bm r} \in \mathbb{R}$ is a scalar representing a small pre-determined number of pixels, $a({\bm o_{i_r}})$ is the area of $r$-th object in $i$-th image, and $\overline a_\text{max}$ and $\overline a_\text{min}$ represent the maximum and minimum areas of objects in $\mathcal{T}$. An example of SA-DCE can be seen in Fig.~\ref{fig:extended-bbox}. 

This subtle yet impactful modification adds peripheral context that is often missing, especially in small object representations, providing the model with additional spatial cues to help in accurate detection. By extending the context, our model can better differentiate objects from the background, leading to improved performance, particularly in complex scenes. Optimization-based methods inherently lack the flexibility to incorporate such targeted context adjustments, as they are confined to the synthetic data representation derived from iterative pixel tuning.

\noindent{\textbf{Objective.}}
We combine the two metrics in Eq.~\ref{ID} and Eq.~\ref{dviersity} into a single objective for data distillation:
\begin{equation}
\mathcal{S}_\mathbf{\hat x}=\argmax_{\mathbf x_T}\ \  \Phi\bigl(\mathbf x_T)+\mathcal{N}\bigl(\mathbf x_T), 
\end{equation}
\vspace{-0.1in}
where $\mathbf x_T$ denotes the final condensed canvas (i.e., synthesized image) after $T$ synthesis iterations.

In practice, we do not explicitly find their optimal values separately, as they are mutually restrictive and entangled. Once the size of the canvas is predefined, we can simply perform an ablation study on {\em the overlap of objects on canvas} for the optimal value that maximizes $\mathcal{S}_\mathbf{\hat x}$, as detailed in the following section.

\noindent During the iterative data-synthesis process, we update $\mathbf x_i$ for $i=0, 1, \ldots, T-1$ using:
\begin{equation}
\mathbf x_{i+1}=f_{\mathrm{remove}} \bigl(f_{\mathrm{add}}(\mathbf x_i)\bigr), 
\quad i \in {0, 1, 2, \ldots, T-1}.
\end{equation}
Here, $f_{\mathrm{add}}(\cdot)$ adds new candidate objects to the current canvas, $f_{\mathrm{remove}}(\cdot)$ filters out low-confidence or redundant objects, thereby refining the composition of $\mathbf x_i$.

\noindent{\bf Iterative Synthesis Methods.}
We consider two iterative processes for building the final canvas $\mathbf x_T$. 

\paragraph{1. First Form: Add-Only.}
The process of this startegy is:
\begin{equation}
\mathbf x_{i+1}=f_{\mathrm{add}}\bigl(\mathbf x_i\bigr), \quad i=0, 1, \dots, T-1, 
\end{equation}
In this scenario, newly added objects remain on the canvas even if their confidence is low and if they overlap other objects smaller than $\tau$. The final objective value is

\vspace{-0.1in}
\begin{equation}
G_1=\Phi\bigl(\mathbf x^{a}_T\bigr)+\mathcal{N}\bigl(\mathbf x^a_T\bigr).
\end{equation}
\paragraph{2. Second Form: Add‐Then‐Remove.}
The {\em Add-Then-Remove} is a loop to construct then refine distilled images:
\begin{equation}
\mathbf x_{i+1}\!=\!f_{\mathrm{remove}}\bigl(f_{\mathrm{add}}(\mathbf x_i)\bigr), \quad i=0, 1, \dots, T-1, 
\end{equation}
Here, each iteration first adds new objects, then filters out objects whose confidence $q(o_j)$ is below a threshold $\eta$, or that fail other criteria (e.g., excessive overlap). The final objective value is
\begin{equation}
G_2=\Phi\bigl(\mathbf x^{ar}_T\bigr)+\mathcal{N}\bigl(\mathbf x^{ar}_T\bigr).
\end{equation}
The following theorem states that {\bf incorporating the remove step} will positively increases the objective, under enough iterations and a well‐chosen confidence threshold.
\begin{theorem}
\label{the:main}
(the proof in Appendix~\ref{sec:proof_thoerem}) Under the above definitions, we have

\vspace{-0.1in}
\begin{equation}
G_2 \ge G_1.
\end{equation}
\end{theorem}

\noindent\textbf{Intuition.}
Because adding a removal step $f_{\mathrm{remove}}(\cdot)$ after every object-addition $f_{\mathrm{add}}(\cdot)$ enables a more refined composition of the canvas, the \emph{second form} is guaranteed to achieve at least as high an objective value as the simpler first form (which lacks a removal step).
That is, the second iterative scheme ({\em add‐then‐remove}) achieves an objective value at least as large as the add‐only approach, under typical assumptions on how objects are added or removed.

\noindent{\textit{Sketch Proof.}}
Setup:
$f_{\text {remove }}$ is an operator that detects objects in the canvas $\mathbf x_i$ and removes those with confidence below a threshold $\eta$. Concretely:

{\em Step-1: Detection step.}
Compute $f_\theta\left(\mathbf x_i\right)$, i.e., run a pre-trained detector on the current canvas $\mathbf x_i$.

{\em Step-2: Scoring each object.}
For each object $o_{i_r}$ in  $\mathbf x_i$ (where $r=1, \ldots, K$, $K$ is the number of objects), obtain a confidence score $q\left(o_{i_r}\right)$.

{\em Step-3: Threshold partition (no overlaps assumed).}
Divide the objects into two groups $\mathcal{O}_1$ and $\mathcal{O}_2$, one with a confidence level greater than the threshold $\eta$, and the other with a confidence level less than or equal to the threshold ${\eta}$:
\begin{equation}
% \footnotesize
\begin{aligned}
& \mathcal{O}_1 := \left\{o_{i_0}, o_{i_1}, \ldots, o_{i_M}\right\},  \text { where } q\left(o_{i_0}\right) \leq \cdots \leq q\left(o_{i_M}\right)<\eta \\
& \mathcal{O}_2 := \left\{o_{i_{M+1}}, o_{i_{M+2}}, \ldots, o_{i_K}\right\}, \\& \text { where } \eta \leq q\left(o_{i_{M+1}}\right) \leq q(o_{i_{M+2}}) \leq \cdots \leq q\left(o_{i_K}\right)
\end{aligned} 
\end{equation}

{\em Step-4: Removing low‐confidence objects.}
All objects whose confidence $< \eta$ are discarded. Thus the \emph{information density} on the canvas changes as follows:
$
\frac{\sum_{r=0}^K a\bigl(o_{i_r}\bigr)\, q\bigl(o_{i_r}\bigr)}{\sum_{r=0}^K a\bigl(o_{i_r}\bigr)}
\longrightarrow
\frac{\sum_{r=M+1}^K a\bigl(o_{i_r}\bigr)\, q\bigl(o_{i_r}\bigr)}{\sum_{r=M+1}^K a\bigl(o_{i_r}\bigr)}.
$

{\bf Comparison of Densities.}
To see why the new density (after removal) is generally higher or equal, we can interpret
$\frac{o_{i_k}}{\sum_{r=0}^K a\bigl(o_{i_r}\bigr)}$
as a probability weight, letting $o_{i_k}$ denote the area $\times$ score contribution of object $k$.
Removing those objects whose confidence is below $\eta$ essentially removes low‐quality (score or area) contributions from the numerator, thereby increasing the average or expected confidence.
If $K$ is sufficiently large, we can consider: $\mathbb{E}_r\left[q\left(o_{i_r}\right)\right]$ and $\mathbb{E}_{r \geq M+1}\left[q\left(o_{i_r}\right)\right]$, a standard probabilistic argument shows that the expected confidence of the surviving set
$\bigl\{o_{i_{M+1}}, \dots, o_{i_{K}}\bigr\}$
is at least as high as that of the entire original set. Formally, 
\begin{equation}
    \mathbb{E}_{r \ge M+1}\bigl[q(o_{i_{r}})\bigr]\;\;\ge\;\;\mathbb{E}_{r}\bigl[q(o_{i_{r}})\bigr], 
\end{equation}
which implies $\Phi\left(\mathbf x_T\right) \geq \Phi\left(\mathbf x_{T-1}\right) \geq \cdots \geq \Phi\left(\mathbf x_0\right)$ in the {\em add-then-remove} scheme.

By similar reasoning (via a probabilistic bound on whether the leftover portion remains undetected), one can show that the presence of overlaps does not harm the objective in the {\em add-then-remove} scheme. Hence, 
$
G_2 \geq G_1
$
even when overlaps are considered. More details are in our Appendix. 

\section{Experiments}
\vspace{-0.05in}
\noindent{\bf Experimental Setup.}
We evaluate \textbf{\texttt{OD\textsuperscript{3}}} with compression ratios ranging from $0.25\%$ to $5\%$ for MS COCO \citep{mscoco} and from $0.5\%$ to $2\%$ for PASCAL VOC \citep{pascal}. We set the overlap threshold $\tau$ to $0.6$ in the candidate selection stage, the confidence threshold $\eta$ to $0.2$ in the candidate screening stage, and $\bf M$ to 40. The foreground objects are inserted into the reconstructed images using their ground truth bounding boxes with extended context using SA-DCE. The backgrounds of the reconstructed images are randomly sampled from the respective datasets. Synthesis experiments are conducted on a single 4090 GPU. The canvas sizes used are $484\times578$ for MS COCO and $375\times500$ for PASCAL VOC, which are the average width and height of the respective full training sets. For the post-evaluation stage, we use VOC2007 and VOC2012 train/val splits combined for synthesis and VOC2007 test set for evaluation. We use standard COCO metrics ($mAP$, $mAP_{50}$, and $mAP_{75}$) along with size metrics ($mAP_{s}$, $mAP_{m}$, and $mAP_{l}$) for the COCO dataset. We use Pascal VOC style $mAP$ and $mAP_{50}$ with the area method that uses all points in the precision-recall curve instead of only 11, which provides a more precise evaluation \citep{pascal}. Each experiment in Tables~\ref{tab:comparison-coco} and \ref{tab:comparison-pascal} was run 4 times with different random seeds. We report the mean and standard error of the mean (± SEM) across these runs. Faster R-CNN-50 models are trained for $96$ epochs and the RetinaNet-50 models for $256$ epochs. All post-evaluation experiments are conducted on 2$\times$ 4090 GPUs, which is highly resource-efficient. Our implementation is based on the mmdetection \citep{mmdetction} and mmrazor \citep{mmrazor} frameworks.

\begin{table}[t!]
\centering
\caption{\textbf{Performance Comparison on MS COCO.} The compression ratios range from $0.25\%$ to $1.0\%$. The observer model and the student model are Faster R-CNN-101 and Faster R-CNN-50.}
\vspace{-0.1in}
\resizebox{0.7\linewidth}{!}{%
\begin{tabular}{c|l|c|c|c}

\toprule
\textbf{\texttt{IPD}}                             & \textbf{Method $\downarrow$} & \textbf{mAP (\%)}            & \textbf{mAP$_{50}$ (\%)}    & \textbf{mAP$_{75}$ (\%)}    \\ \midrule
\multirow{6}{*}{0.25\%} & Random     & 3.50   & 9.70  & 1.60 \\
                        & Uniform    & 3.60   & 9.80  & 1.60 \\
                        & K-Center   & 1.70   & 6.30  & 0.40 \\
                        & Herding    & 1.70   & 5.80  & 0.50 \\
                        & DCOD \citep{fetchandforge} & 7.20   & 17.20 & 4.80 \\
                        & \cellcolor{gray!30}\textbf{\texttt{OD\textsuperscript{3}} (Ours)} & \cellcolor{gray!30}\textbf{$\textbf{12.90$\pm$0.1}_{\textbf{(\texttt{+}5.7)}}$} \tikz \fill[green!80!black] (0,0) -- (0.2,0) -- (0.1,0.2) -- cycle; & \cellcolor{gray!30}\textbf{$\textbf{24.30$\pm$0.2}_{\textbf{(\texttt{+}7.1)}}$} \tikz \fill[green!80!black] (0,0) -- (0.2,0) -- (0.1,0.2) -- cycle;& \cellcolor{gray!30}\textbf{$\textbf{12.10$\pm$0.3}_{\textbf{(\texttt{+}7.3)}}$} \tikz \fill[green!80!black] (0,0) -- (0.2,0) -- (0.1,0.2) -- cycle;\\ \midrule
\multirow{6}{*}{0.5\%}  & Random     & 5.50   & 14.20 & 2.90 \\
                        & Uniform    & 5.60   & 14.30 & 2.90 \\
                        & K-Center   & 2.80   & 8.90  & 0.70 \\
                        & Herding    & 2.60   & 8.80  & 0.80 \\
                        & DCOD \citep{fetchandforge} & 10.00  & 21.50 & 8.00 \\
                        & \cellcolor{gray!30}\textbf{\texttt{OD\textsuperscript{3}} (Ours)} & \cellcolor{gray!30}\textbf{$\textbf{17.20$\pm$0.1}_{\textbf{(\texttt{+}7.2)}}$} \tikz \fill[green!80!black] (0,0) -- (0.2,0) -- (0.1,0.2) -- cycle; & \cellcolor{gray!30}\textbf{$\textbf{31.90$\pm$0.2}_{\textbf{(\texttt{+}10.4)}}$} \tikz \fill[green!80!black] (0,0) -- (0.2,0) -- (0.1,0.2) -- cycle; & \cellcolor{gray!30}\textbf{$\textbf{16.90$\pm$0.1}_{\textbf{(\texttt{+}7.9)}}$} \tikz \fill[green!80!black] (0,0) -- (0.2,0) -- (0.1,0.2) -- cycle; \\ \midrule
\multirow{6}{*}{1.0\%}    & Random     & 8.30   & 19.70 & 5.30 \\
                        & Uniform    & 8.40   & 19.70 & 5.40 \\
                        & K-Center   & 4.00   & 12.90 & 1.20 \\
                        & Herding    & 4.10   & 12.50 & 1.30 \\
                        & DCOD \citep{fetchandforge} & 12.10  & 24.70 & 10.40 \\
                        & \cellcolor{gray!30}\textbf{\texttt{OD\textsuperscript{3}} (Ours)} & \cellcolor{gray!30}\textbf{$\textbf{22.40$\pm$0.1}_{\textbf{(\texttt{+}10.3)}}$} \tikz \fill[green!80!black] (0,0) -- (0.2,0) -- (0.1,0.2) -- cycle; & \cellcolor{gray!30}\textbf{$\textbf{39.50$\pm$0.2}_{\textbf{(\texttt{+}14.8)}}$} \tikz \fill[green!80!black] (0,0) -- (0.2,0) -- (0.1,0.2) -- cycle;& \cellcolor{gray!30}\textbf{$\textbf{22.90$\pm$0.1}_{\textbf{(\texttt{+}12.5)}}$} \tikz \fill[green!80!black] (0,0) -- (0.2,0) -- (0.1,0.2) -- cycle;\\ \midrule
                        \multicolumn{2}{c|}{\textbf{Whole Dataset}}& 39.80 & 60.10  & 43.30  \\ \bottomrule
\end{tabular}%
}
\label{tab:comparison-coco}
\vspace{-0.05in}
\end{table}

\begin{table}[t!]
\centering
\caption{\textbf{Performance Comparison on Pascal VOC (mAP$_{50}$\%).} The compression ratios (IPD) range from $0.5\%$ to $2.0\%$. The observer and target model are both Faster R-CNN50.}
\label{tab:comparison-pascal}
\vspace{-0.1in}
\resizebox{0.8\linewidth}{!}{%
\begin{tabular}{c|c|c|c|c|c|c}
\toprule
\textbf{IPD} & \textbf{Random} & \textbf{Herding} & \textbf{K-center} & \textbf{Uniform} & \textbf{DCOD \citep{fetchandforge}} & \cellcolor{gray!30}\textbf{\texttt{OD\textsuperscript{3}} (Ours)} \\
\midrule
0.5\% & 15.80 & 12.60 & 14.50 & 15.80 & 37.90 & \cellcolor{gray!30}\textbf{$\textbf{38.50$\pm$0.1}_{\textbf{(\texttt{+}0.6)}}$} \tikz \fill[green!80!black] (0,0) -- (0.2,0) -- (0.1,0.2) -- cycle; \\
1.0\% & 25.50 & 19.30 & 21.90 & 25.70 & 46.40 & \cellcolor{gray!30}\textbf{$\textbf{51.10$\pm$0.2}_{\textbf{(\texttt{+}4.7)}}$} \tikz \fill[green!80!black] (0,0) -- (0.2,0) -- (0.1,0.2) -- cycle; \\
2.0\% & 40.50 & 28.10 & 31.30 & 40.60 & 50.70 & \cellcolor{gray!30}\textbf{$\textbf{58.70$\pm$0.1}_{\textbf{(\texttt{+}8.0)}}$} \tikz \fill[green!80!black] (0,0) -- (0.2,0) -- (0.1,0.2) -- cycle;\\
\midrule
\textbf{Whole Dataset} & \multicolumn{6}{c}{80.35} \\
\bottomrule
\end{tabular}
}
\vspace{-0.15in}
\end{table}

\begin{table}[]
\centering
\vspace{-0.15in}
\caption{\textbf{Ablation Study on Label Type.} The impact of using mask labels, bounding box (BBox) labels, or Ex-BBox in the data synthesis process across various compression ratios. {\em Ex-Bbox} represents the BBox with the extended context using SA-DCE.}
\vspace{-0.1in}
\resizebox{0.9\linewidth}{!}{%
\begin{tabular}{c|l|cccccc}
\toprule
\textbf{\texttt{IPD}}                            & \textbf{Label} & \textbf{mAP}            & \textbf{mAP$_{50}$}    & \textbf{mAP$_{75}$}    & \textbf{mAP$_s$}       & \textbf{mAP$_m$}       & \textbf{mAP$_l$ }      \\ \midrule
\multirow{3}{*}{\textbf{0.25\%}} 
& Mask  & 
10.90&20.80 &10.30 &3.50 &15.10 &15.90 \\
& Bbox  &12.40&23.90 &11.60 &4.90 &16.10&\textbf{18.10}\\
& \cellcolor{gray!30}\textbf{Ex-Bbox} &\cellcolor{gray!30}\textbf{$\textbf{12.90}_{\textbf{(\texttt{+}0.5)}}$}\tikz \fill[green!80!black] (0,0) -- (0.2,0) -- (0.1,0.2) -- cycle; &\cellcolor{gray!30}\textbf{$\textbf{24.30}_{\textbf{(\texttt{+}0.4)}}$}\tikz \fill[green!80!black] (0,0) -- (0.2,0) -- (0.1,0.2) -- cycle; 
&\cellcolor{gray!30}\textbf{$\textbf{12.10}_{\textbf{(\texttt{+}0.7)}}$}\tikz \fill[green!80!black] (0,0) -- (0.2,0) -- (0.1,0.2) -- cycle; &\cellcolor{gray!30}\textbf{$\textbf{5.60}_{\textbf{(\texttt{+}0.7)}}$}\tikz \fill[green!80!black] (0,0) -- (0.2,0) -- (0.1,0.2) -- cycle; 
&\cellcolor{gray!30}\textbf{$\textbf{16.80}_{\textbf{(\texttt{+}0.7)}}$}\tikz \fill[green!80!black] (0,0) -- (0.2,0) -- (0.1,0.2) -- cycle; 
&\cellcolor{gray!30}\textbf{$\textbf{17.70}_{\textbf{(\texttt{-}0.4)}}$}\tikz \fill[red!90!black] (0,0.2) -- (0.2,0.2) -- (0.1,0) -- cycle;\\
\midrule

\multirow{3}{*}{\textbf{0.5\%}} 
& Mask  & 15.10 & 28.00 & 14.80& 5.60 & 20.30 &22.30\\
& Bbox & 16.60& 30.30&16.30&6.80& 21.70&\textbf{23.30}\\

& \cellcolor{gray!30}\textbf{Ex-Bbox} 

&\cellcolor{gray!30}\textbf{$\textbf{17.20}_{\textbf{(\texttt{+}0.6)}}$}\tikz \fill[green!80!black] (0,0) -- (0.2,0) -- (0.1,0.2) -- cycle;

&\cellcolor{gray!30}\textbf{$\textbf{31.90}_{\textbf{(\texttt{+}1.6)}}$}\tikz \fill[green!80!black] (0,0) -- (0.2,0) -- (0.1,0.2) -- cycle;

&\cellcolor{gray!30}\textbf{$\textbf{16.90}_{\textbf{(\texttt{+}0.6)}}$}\tikz \fill[green!80!black] (0,0) -- (0.2,0) -- (0.1,0.2) -- cycle;

&\cellcolor{gray!30}\textbf{$\textbf{8.40}_{\textbf{(\texttt{+}1.6)}}$}\tikz \fill[green!80!black] (0,0) -- (0.2,0) -- (0.1,0.2) -- cycle;

&\cellcolor{gray!30}\textbf{$\textbf{23.00}_{\textbf{(\texttt{+}1.3)}}$}\tikz \fill[green!80!black] (0,0) -- (0.2,0) -- (0.1,0.2) -- cycle;

&\cellcolor{gray!30}\textbf{$\textbf{22.80}_{\textbf{(\texttt{-}0.5)}}$}\tikz \fill[red!90!black] (0,0.2) -- (0.2,0.2) -- (0.1,0) -- cycle;
\\ 
\midrule

\multirow{3}{*}{\textbf{1.0\%}} 
& Mask&21.20 & 37.40 & 21.30 & 8.90 &26.40 &29.90 \\
& Bbox  & 22.00 &38.70 &22.30&9.70&27.40&\textbf{30.10}\\
& \cellcolor{gray!30}\textbf{Ex-Bbox} &\cellcolor{gray!30}\textbf{$\textbf{22.40}_{\textbf{(\texttt{+}0.4)}}$}\tikz \fill[green!80!black] (0,0) -- (0.2,0) -- (0.1,0.2) -- cycle; &\cellcolor{gray!30}\textbf{$\textbf{39.50}_{\textbf{(\texttt{+}0.8)}}$}\tikz \fill[green!80!black] (0,0) -- (0.2,0) -- (0.1,0.2) -- cycle;
&\cellcolor{gray!30}\textbf{$\textbf{22.90}_{\textbf{(\texttt{+}0.6)}}$}\tikz \fill[green!80!black] (0,0) -- (0.2,0) -- (0.1,0.2) -- cycle;
&\cellcolor{gray!30}\textbf{$\textbf{10.60}_{\textbf{(\texttt{+}0.9)}}$}\tikz \fill[green!80!black] (0,0) -- (0.2,0) -- (0.1,0.2) -- cycle;
&\cellcolor{gray!30}\textbf{$\textbf{28.00}_{\textbf{(\texttt{+}0.6)}}$}\tikz \fill[green!80!black] (0,0) -- (0.2,0) -- (0.1,0.2) -- cycle;
&\cellcolor{gray!30}\textbf{$\textbf{29.80}_{\textbf{(\texttt{-}0.3)}}$}\tikz \fill[red!90!black] (0,0.2) -- (0.2,0.2) -- (0.1,0) -- cycle;\\ 
\midrule

\multirow{3}{*}{\textbf{5.0\%}} 
& Mask  & 30.00 & 49.50 & 31.70 & 15.10& 34.80& \textbf{39.10}\\
& Bbox  &29.90 &49.40&31.50 &15.00 &34.70& 38.50\\

& \cellcolor{gray!30}\textbf{Ex-Bbox} 

&\cellcolor{gray!30}\cellcolor{gray!30}\textbf{$\textbf{30.10}_{\textbf{(\texttt{+}0.2)}}$}\tikz \fill[green!80!black] (0,0) -- (0.2,0) -- (0.1,0.2) -- cycle;
&\cellcolor{gray!30}\textbf{$\textbf{49.70}_{\textbf{(\texttt{+}0.3)}}$}\tikz \fill[green!80!black] (0,0) -- (0.2,0) -- (0.1,0.2) -- cycle;
&\cellcolor{gray!30}\textbf{$\textbf{31.80}_{\textbf{(\texttt{+}0.3)}}$}\tikz \fill[green!80!black] (0,0) -- (0.2,0) -- (0.1,0.2) -- cycle;
&\cellcolor{gray!30}\textbf{$\textbf{16.20}_{\textbf{(\texttt{+}1.2)}}$}\tikz \fill[green!80!black] (0,0) -- (0.2,0) -- (0.1,0.2) -- cycle;
&\cellcolor{gray!30}\textbf{$\textbf{34.90}_{\textbf{(\texttt{+}0.2)}}$}\tikz \fill[green!80!black] (0,0) -- (0.2,0) -- (0.1,0.2) -- cycle;
&\cellcolor{gray!30}\textbf{$\textbf{38.40}_{\textbf{(\texttt{-}0.1)}}$}\tikz \fill[red!90!black] (0,0.2) -- (0.2,0.2) -- (0.1,0) -- cycle;\\ 
\bottomrule
\end{tabular}%
}

\label{tab:bboxmask}
\end{table}

\begin{table}[t!]
% \vspace{-0.18in}
\caption{\textbf{Cross-Architecture Performance Comparison for \texttt{OD\textsuperscript{3}} on MS COCO.} Observer models use ResNet101 and target models use ResNet50.}
\vspace{-0.1in}
\centering
\resizebox{0.8\linewidth}{!}{%
\begin{tabular}{c|c|c|c|c|c}
\toprule
\textbf{\texttt{IPD}}        & \textbf{Observer}   & \textbf{Target}       & \textbf{mAP (\%)}            & \textbf{mAP$_{50}$ (\%)}    & \textbf{mAP$_{75}$ (\%)}    \\ \midrule
\multirow{4}{*}{\textbf{0.25\%}}
& RetinaNet  & RetinaNet & 13.90&25.30&13.50\\
& Faster R-CNN   &  RetinaNet& 14.50 &25.70 &14.30 \\
& Deformable DETR & Faster R-CNN & 11.90 & 22.60 & 11.10 \\
& ViTDet & Faster R-CNN &11.00 & 21.30 & 	10.10 \\
                
 \midrule
\multirow{4}{*}{\textbf{0.5\%}}
& RetinaNet&RetinaNet&18.40&32.50&18.20 \\
& Faster R-CNN&  RetinaNet    &17.40 &30.20 &17.60  \\

& Deformable DETR & Faster R-CNN & 	16.20 & 29.50 &	16.00 \\
& ViTDet & Faster R-CNN & 15.90 & 29.20 &15.60 \\
\midrule

\multirow{4}{*}{\textbf{1.0\%}}
& RetinaNet  & RetinaNet &22.20&37.90&22.60\\
& Faster R-CNN&  RetinaNet& 22.20 & 37.40 & 23.00\\

& Deformable DETR & Faster R-CNN & 22.00 & 38.00 &	22.90 \\
& ViTDet  & Faster R-CNN &21.70&38.30&21.80\\ 

\bottomrule
\end{tabular}%
}
\vspace{-0.1in}
\label{tab:cross_architecture}
\end{table}

\begin{table}[ht]
\centering
\caption{\textbf{Ablation Study on Method Components.} We highlight the impact of candidate selection and screening on MS COCO performance across varying compression rates.}
\vspace{-0.1in}
\resizebox{0.8\linewidth}{!}{%
\begin{tabular}{>{\centering\arraybackslash}m{1.5cm}|cc|cccccc}
\toprule
\textbf{\texttt{IPD}} & \makecell{\textbf{Candidate} \\ \textbf{Selection}} &\makecell{\textbf{Candidate} \\ \textbf{Screening}} & \textbf{mAP}            & \textbf{mAP$_{50}$}    & \textbf{mAP$_{75}$}    & \textbf{mAP$_s$}       & \textbf{mAP$_m$}       & \textbf{mAP$_l$ }      \\ \midrule
\multirow{3}{*}{\textbf{0.25\%}} 
& \xmark& \xmark &0.90&2.40&0.40&0.00&1.10&1.20
   \\ 
& \checkmark & \xmark & 9.70 &19.10 &8.90&3.70 &13.10&13.60 \\
&\checkmark&\checkmark&\textbf{12.90}&\textbf{24.30} &\textbf{12.10} &\textbf{5.60} &\textbf{16.80}&\textbf{17.70} \\ \midrule

\multirow{3}{*}{\textbf{0.5\%}}  
& \xmark& \xmark& 2.00&4.00&1.90 &0.10&2.60 &3.10 \\ 
& \checkmark & \xmark &14.50&27.30&13.80&5.90&19.30&20.30\\
& \checkmark & \checkmark & \textbf{17.20}& \textbf{31.90}&\textbf{16.90}&\textbf{8.40}&
\textbf{23.00}&\textbf{22.80}\\ \midrule

\multirow{3}{*}{\textbf{1.0\%}}    
& \xmark  & \xmark &7.50&14.10&7.30&0.9&8.90 &13.30 \\ 
& \checkmark & \xmark&19.00& 33.90& 19.10& 8.70&24.40 &25.70 
   \\ 
& \checkmark & \checkmark &\textbf{22.40} &\textbf{39.50} &\textbf{22.90}&\textbf{10.60}&\textbf{28.00}&\textbf{29.80}\\
\midrule

\multirow{3}{*}{\textbf{5.0\%}}    
& \xmark                         & \xmark&8.60&17.80&7.30&1.10& 10.40&16.40  \\ 
& \checkmark & \xmark& 28.10 &46.70& 29.60& 14.00& 34.00 & 37.00   \\
& \checkmark & \checkmark &\textbf{30.10}&\textbf{49.70}&\textbf{31.80}&\textbf{16.20}&\textbf{34.90}&\textbf{38.40}  \\ \bottomrule
\end{tabular}%
}
\label{tab:selection_screening}
\vspace{-0.1in}
\end{table}

\begin{figure*}[b]
    \centering
    % Column 1
    \vspace{-0.1in}
    \begin{minipage}[t]{0.9\textwidth}
        \centering
        \vspace{-0.15in}
        \includegraphics[width=0.23\textwidth, height=\textheight, keepaspectratio]{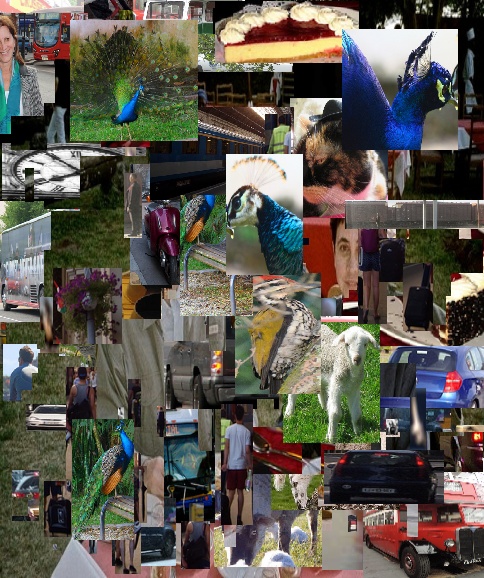}
        \hfill
        \includegraphics[width=0.23\textwidth, height=\textheight, keepaspectratio]{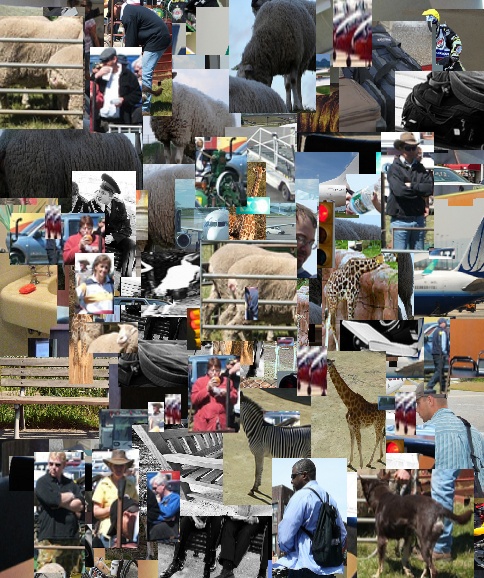}
        \hfill
        \includegraphics[width=0.23\textwidth, height=\textheight, keepaspectratio]{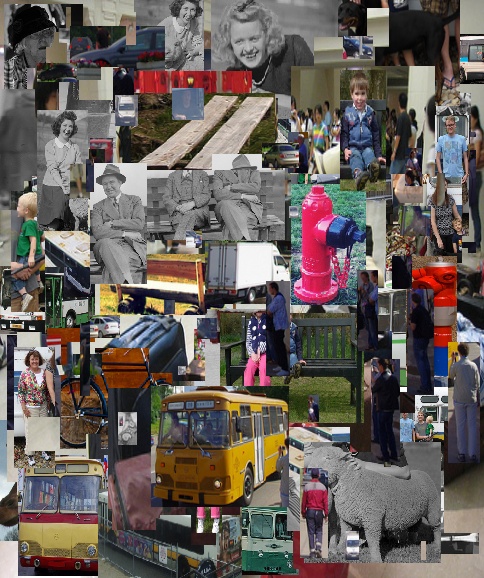}
        \hfill
        \includegraphics[width=0.23\textwidth, height=\textheight, keepaspectratio]{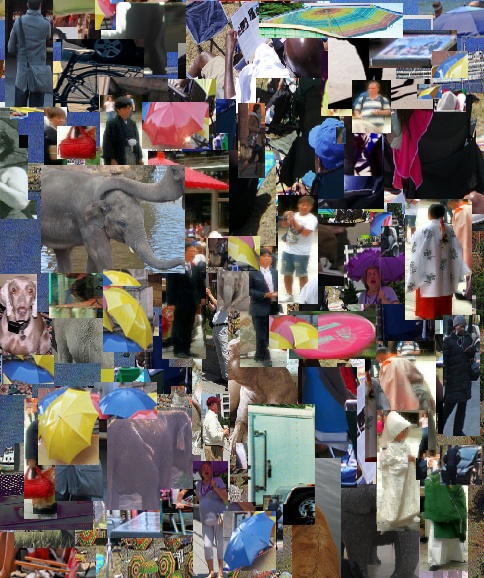}
    \end{minipage}
    \vspace{-0.1in}
    \caption{\textbf{Qualitative results of the synthesis process of \textbf{\texttt{OD\textsuperscript{3}}} on MS COCO}. Initial backgrounds of the canvas are randomly selected from the training set, and objects are inserted using their bounding-box level labels. Those images are generated at $0.5\%$ \texttt{IPD}.}
    \label{fig:qualitative}
\end{figure*}

\noindent{\bf Image Generation Time and Efficiency.} Our synthesis process is highly efficient compared to optimization-based approaches like DCOD (which did not report generation time). Our primary time overhead comes from observer screening. Specifically, generating the condensed dataset takes approximately 4.7 hours on MS COCO and 0.74 hours on PASCAL VOC using a single 4090 GPU.

\vspace{-0.1in}
\subsection{Experimental Results}
\vspace{-0.05in}
Table \ref{tab:comparison-coco} presents the comparative results of our method on MS COCO \citep{mscoco} with core-set selection methods and with DCOD \citep{fetchandforge}. The core-set selection methods include: random initialization \citep{rebuffi2017icarl}, Uniform \citep{lee2024coreset}, K-center \citep{sener2017active}, and Herding \citep{castro2018end, chen2012super}. Our method, \textbf{\texttt{OD\textsuperscript{3}}}, outperforms all other methods across various compression ratios (\texttt{IPD}) ranging from 0.25\% to 1.0\%. Notably, at 1.0\%, we achieve a substantial \textbf{$14.8\%$} improvement in $mAP_{50}$ over DCOD. Furthermore, our method consistently outperforms other core-set selection methods, with $mAP_{50}$ improvements of up to 27.0\% at 1.0\% \texttt{IPD}. Our method also achieves the highest performance in $mAP$, $mAP_{50}$ and $mAP_{75}$ at each compression ratio. Since the authors of DCOD did not report its performance on the size metrics of MS COCO, we are unable to compare the methods in that regard. Nonetheless, these results underline the effectiveness of \textbf{\texttt{OD\textsuperscript{3}}} in achieving superior performance across a range of compression ratios. We also report results on PASCAL VOC in Table~\ref{tab:comparison-pascal} with different \texttt{IPDs}, where the method achieves 58.70\% $mAP$ at 2.0\% compression, surpassing DCOD by 8.0\%. Qualitative visualizations of the synthesis from our \textbf{\texttt{OD\textsuperscript{3}}} on MS COCO are shown in Fig.~\ref{fig:qualitative}.

\vspace{-0.1in}
\subsection{Ablation Studies}
\vspace{-0.05in}

\begin{table}[t!]
\centering
\caption{\textbf{Performance across higher compression ratios on MS COCO.} While our main experiments evaluate extreme compression regimes (0.25\%–1.0\% IPD), this table reports performance at larger compression ratios (5.0\%–20.0\%), demonstrating that \texttt{OD\textsuperscript{3}} continues to scale predictably as more synthesized data are allocated.}
\vspace{-0.1in}
\resizebox{0.85\linewidth}{!}{%
{
\begin{tabular}{c|c|c|c|c|c|c}
\toprule
\textbf{\texttt{IPD}} & \textbf{mAP (\%)} & \textbf{mAP$_{50}$ (\%)} & \textbf{mAP$_{75}$ (\%)} & \textbf{mAP$_s$ (\%)} & \textbf{mAP$_m$ (\%)} & \textbf{mAP$_l$ (\%)} \\ \midrule

\textbf{5.0\%}  & 30.10 & 49.70 & 31.80 & 16.20 & 34.90 & 38.40 \\ 
\textbf{10.0\%} & 31.80 & 51.70 & 34.10 & 16.80 & 36.70 & 40.50 \\ 
\textbf{20.0\%} & 32.60 & 53.00 & 34.80 & 18.00 & 37.50 & 41.60 \\ \midrule

\multicolumn{1}{c|}{\textbf{Whole Dataset}} &
39.80 &
60.60 &
43.50 &
22.40 &
43.90 &
51.90 \\

\bottomrule
\end{tabular}
}}
\label{tab:scalability}
\vspace{-0.15in}
\end{table}

\textbf{Label type.} The type of label used when inserting the candidate objects into the synthesized image is studied in Table~\ref{tab:bboxmask}. We consider three types of labels: mask-level label, BBox-level label, and \textit{Ex-Bbox}, which refers to a BBox with extended context using SA-DCE (refer to Sec.~\ref{subsec:framework}). Using BBox labels outperforms the mask labels across all \texttt{IPDs} except for $5.0\%$, where their performance converges to a similar level. This is because Bbox labels preserves local context and surrounding environment of individual objects, providing models with additional cues for recognizing objects. Using \textit{Ex-Bbox} further improves performance across all \texttt{IPDs}, where an improvement of $1.8\%$ in $mAP_{50}$ can be seen at $0.5\%$ compression. When specifically analyzing the size metrics, the extended context benefits small objects the most on the account of large objects, which bridges the substantial gap between their detection performance.

\begin{wrapfigure}{r}{0.6\textwidth}
    \centering
    \vspace{-0.15in}
    \includegraphics[width=\linewidth]{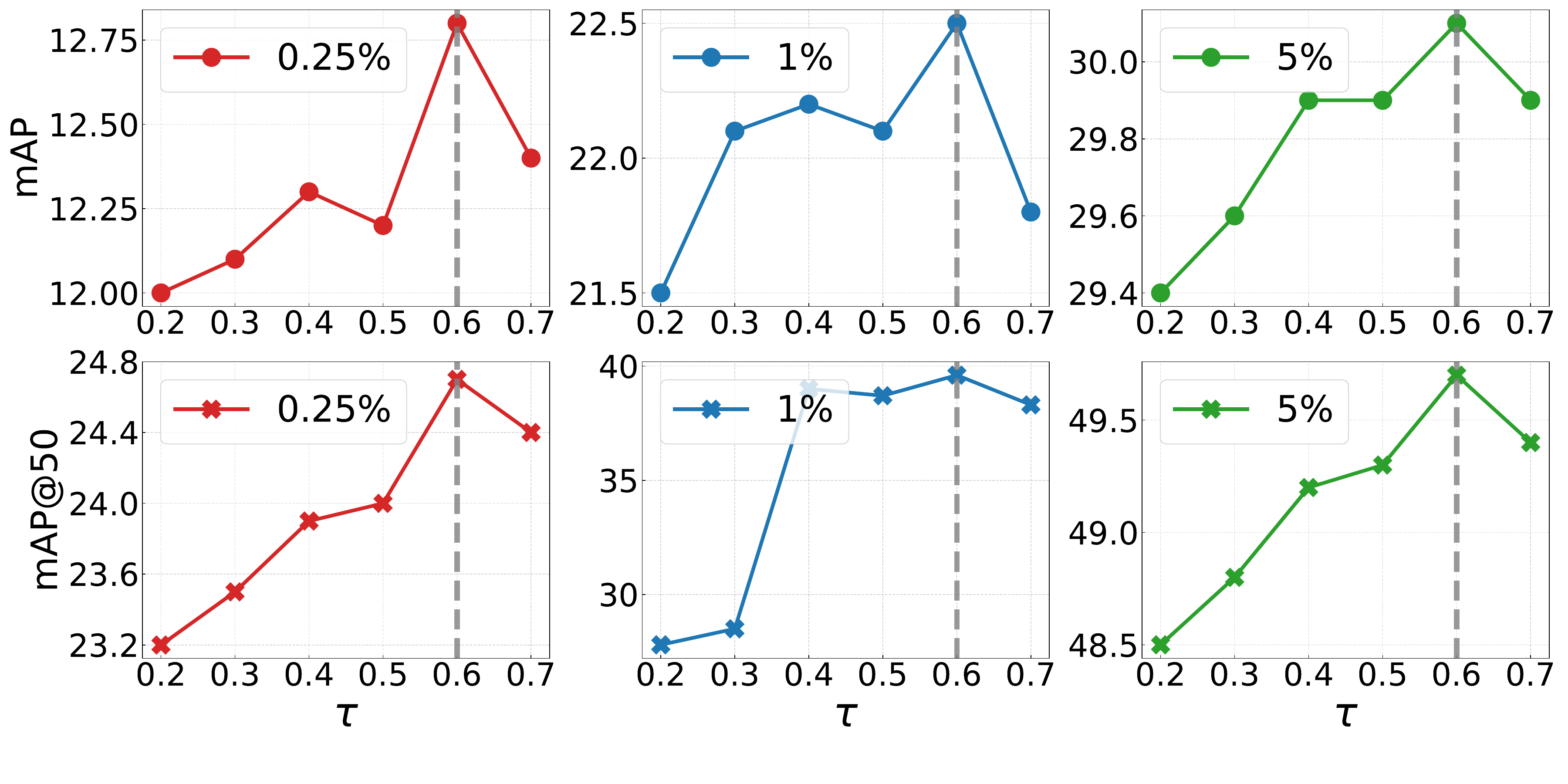}
    \vspace{-0.35in}
    \caption{\textbf{Ablation study of overlap threshold $\bm \tau$.} mAP and mAP$_{50}$ are evaluated at different thresholds used in candidate selection with compression ratios ranging from 0.25\% to 5\%.} 
    \label{fig:tau-ablation}
    \vspace{-0.17in}
\end{wrapfigure}

\textbf{Overlap threshold.} Fig.~\ref{fig:tau-ablation} shows how different values of overlap thresholds $\tau$ in candidate selection affect the performance of our method across various compression ratios. It can be seen that $0.6$ consistently optimizes $mAP_{50}$ and $mAP$. This value can be thought of as an optimal trade-off between $\Phi(\mathbf x)$ and $\mathcal{N}(\mathbf x)$.

\textbf{Cross-architecture generalization.} To assess the generalization capability of our condensed datasets, we conduct experiments with Faster R-CNN \citep{ren2015faster}, a two-stage detector and RetinaNet \citep{lin2017focal}, a one-stage detector. Table \ref{tab:cross_architecture} shows that the distilled datasets can generalize in heterogeneous settings, where the observer model is a two-stage detector, and the target model is a one-stage detector, across varying compression ratios from 0.25\% to 5.0\%. The results demonstrate that performance on RetinaNet is comparable to that on Faster R-CNN across all \texttt{IPDs}. Specifically, at 0.25\% \texttt{IPD}, $mAP_{50}$ for the Faster R-CNN observer and RetinaNet target configuration reaches 25.70\%, surpassing the 24.30\% obtained in the Faster R-CNN observer and target setup. At higher compression ratios, such as 1.0\% and 5.0\%, RetinaNet continues to yield competitive results, achieving $mAP_{50}$ scores of 37.40\% and 48.60\%, respectively. In addition, we evaluate ViTDet \citep{li2022exploring} and Deformable DETR \citep{zhu2020deformable}, transformer-based detectors, as observer models. For instance, ViTDet achieves strong performance with Faster R-CNN as a target model, reaching $mAP_{50}$ scores of 21.30\%, 29.20\%, and 38.30\% at 0.25\%, 0.5\%, and 1.0\%  \texttt{IPD}, respectively, despite their architectural differences. The results demonstrate that our method maintains transferability across fundamentally different model paradigms, further highlighting the robustness of the distilled datasets and their effective applicability across diverse architectures.

\textbf{Method Components.} Table \ref{tab:selection_screening} presents an ablation study evaluating the impact of candidate selection and candidate screening on the MS COCO performance across varying compression ratios (\texttt{IPD}). The table entries where both are not used correspond to when all objects from the training set are randomly assigned a location and inserted into the distilled images without any filtration. The results demonstrate the effectiveness of both components in improving the quality of the synthesized dataset. The addition of candidate screening further improves the results across all compression ratios. For example, there is $3.4\%$ and a $5.6\%$ increase in $mAP$ and $mAP_{50}$ for the $1.0\%$ distilled dataset.

\textbf{More IPD budgets}. Table~\ref{tab:scalability} summarizes performance when training on distilled datasets and the IPD budget is increased beyond the extreme compression regimes used in our main experiments (0.25\% to 1.0\%). As more synthesized data are allocated (5\%–20\% IPD), performance rises steadily across all COCO metrics, indicating that \texttt{OD\textsuperscript{3}} scales in a predictable and stable manner. Although these larger budgets approach the performance of the full dataset, a consistent gap remains, which reflects the intrinsic challenge of reproducing the full diversity of COCO using synthesized samples. 

\section{Conclusion}

In this work, we introduced a new \textbf{\texttt{OD\textsuperscript{3}}} framework for optimization-free dataset distillation in object detection, achieving significant improvements over existing methods. Using a novel two-stage process of \textit{candidate selection} and \textit{candidate screening} driven by a pre-trained observer model, our framework strategically synthesizes compact yet highly effective datasets tailored for object detection. Our method consistently demonstrated superior performance across multiple evaluation metrics. For instance, 
\textbf{\texttt{OD\textsuperscript{3}}} achieved more than \textbf{$14.0\%$} improvement in mAP$_{50}$ compared to the state-of-the-art method DCOD on MS COCO.

\section*{Ethics Statement}
This work uses only publicly available, appropriately licensed detection datasets (e.g., COCO) and does not involve new data collection or identity inference. Distilled data can inherit and potentially amplify biases present in the sources, we therefore report detailed metrics and provide extensive distilled visualizations to monitor this. The method is not intended for surveillance or biometric identification. To reduce environmental impact, our optimization-free distillation aims to lower compute and storage costs. 

\section*{Reproducibility Statement}

For reproducibility, we will release: (i) code with exact scripts and command lines for generating distilled sets and training/evaluating detectors, (ii) pinned dependencies and a conda environment file, (iii) random seeds and deterministic flags, (iv) the distilled datasets themselves, and (v) details reproducing all tables/figures. We detail data splits, preprocessing, evaluation and report mean, $\pm$, std over multiple runs. Ablation scripts cover distilled size and cross-architecture transfer (e.g., Faster R-CNN, RetinaNet, etc.).

\bibliography{bibliography}
\bibliographystyle{iclr2026_conference}

%%%%%%%%%%%%%%%%%%%%%%%%%%%%%%%%%%%%%%%%%%%%%%%%%%%%%%%%%%%%

\appendix

\newpage

\section*{\Large{Appendix}}

\section{Limitations and Societal Impacts}
\label{append:limitations}
There are many potential societal impacts of our work, such as improving the accessibility of efficient datasets for academia with limited computational resources, and fostering the development of sustainable AI. While our method does not introduce direct negative consequences, it is important to acknowledge that object detection technology can be misused - particularly in surveillance applications that infringe on individual privacy. The increased efficiency and scalability enabled by dataset distillation may unintentionally lower barriers for deploying such systems at scale. One limitation of our approach is the reliance on ground-truth bounding boxes during synthesis, which assumes access to labeled data. This restricts the method's applicability in fully unsupervised or label-scarce scenarios.

\section{Algorithm}

Our detailed procedure is shown in Algorithm~\ref{alg:od3}. First, each object candidate is added to the partially formed ``blank canvas'' via \texttt{random copy-paste}. Multiple objects may be overlaid, so that visual variety is preserved. Next, the observer model runs on this synthesized image and assesses the confidence of each placed object. Low-confidence objects that are not matched to the ground truth objects are removed, refining the canvas into a more coherent scene. This cycle of {\em add-and-remove} iterates multiple times, driving the canvas toward a final state containing only high-confidence, mid-overlapping objects. Fig.~\ref{fig:framework} green boxes in screening stage indicate an inserted object is deemed infeasible, applying removal process to maintain quality and coherence.

\RestyleAlgo{ruled}
\SetKwComment{Comment}{$\triangleright$\ }{}
\SetKwInput{KwInput}{Input}
\SetKwInput{KwOutput}{Output}
\SetKwBlock{KwStage}{Stage}{}

\begin{algorithm}[h]
\caption{\textbf{O}ptimization-free \textbf{D}ataset \textbf{D}istillation for Object \textbf{D}etection (\textbf{\texttt{OD\textsuperscript{3}}})}\label{alg:od3}
\KwInput{
     Original dataset $\mathcal{T}$; Synthetic dataset $\mathcal{S}$; Observer model $\theta_{\text{obs}}$; Overlap threshold $\tau$; Screening threshold $\eta$; Images per dataset \texttt{IPD}; Canvas $\mathcal{C}$ (initially $\varnothing$ and updated constantly with $\mathbf{\hat{x}}$); Extension $\bm \ell$ in Eq.~\ref{extension}; Random placement candidate positions $\langle \mathbf{m}_t, \mathbf{n}_t \rangle$ for $t = 1, \dots, \bf M$.
} 

\For{$\mathbf{\hat x}_j \in \mathcal{S}$ where $|\mathcal{S}| = {\texttt{\em IPD}}$}{
 \While{$\mathcal{C}$ is not full}{  %There is still space on the canvas to insert object.
    \Comment{Candidate Selection \& Placement}
    \For{$(\mathbf x_{i}, \mathbf y_{i}) \in \mathcal{T}$}
    {  
        \For{$<\!\bm b_{ir}, \bm c_{ir}\!>\in \mathbf y_{i}$}{
                \ \ $\bm b'_{ir} \leftarrow \bm b_{ir} + \bm \ell_{ir}$ \Comment{$\bm \ell_{ir} \leftarrow F(\overline r)$\;}
                \vspace{-0.08in}
                \While{$\text{IoU}(\bm b'_{ir}, \langle \mathbf{m}_t, \mathbf{n}_t\rangle, \mathcal{C}) < \tau$  \textbf{and} $\text{attempts} < \bf M$}{
                    Place $<\!\bm b'_{ir}, \bm c_{ir}\!>$ $\rightarrow$ $\mathcal{C}$; Exit\;
                }
        }
    }
    \Comment{Candidate Screening}
    Filter objects from $\mathcal{C}$ for $\mathbf{\hat{x}}$\;
    $\mathbf{\hat{y}}_{\text{obs}} = \theta_{\text{obs}}(\mathcal{{C}})$\;
    \For{$<\!\bm b_{k}, \bm c_{k}\!> \in \mathcal{C}$}
    {   
        \If{$\text{Conf}(\mathbf{\hat{y}}_{\text{obs}}, \bm b_{k}) < \eta$}
        {
            Remove $<\!\bm b_{k}, \bm c_{k}\!>$ from $\mathcal{C}$\;
        }
    }
    $\mathbf{\hat{x}}_j\leftarrow\mathcal{C}$\;
   }
   Append $\mathbf{\hat{x}}_j$  to $\mathcal{S}$\;
}
\KwOutput{Synthesized dataset $\mathcal{S}$}
\end{algorithm}

\clearpage
\section{Distilled Data Distribution Statistics}

\begin{wrapfigure}{r}{0.5\textwidth}
  \begin{center}
  \vspace{-\intextsep}
  \vspace{-0.05in}
\includegraphics[width=0.5\textwidth]{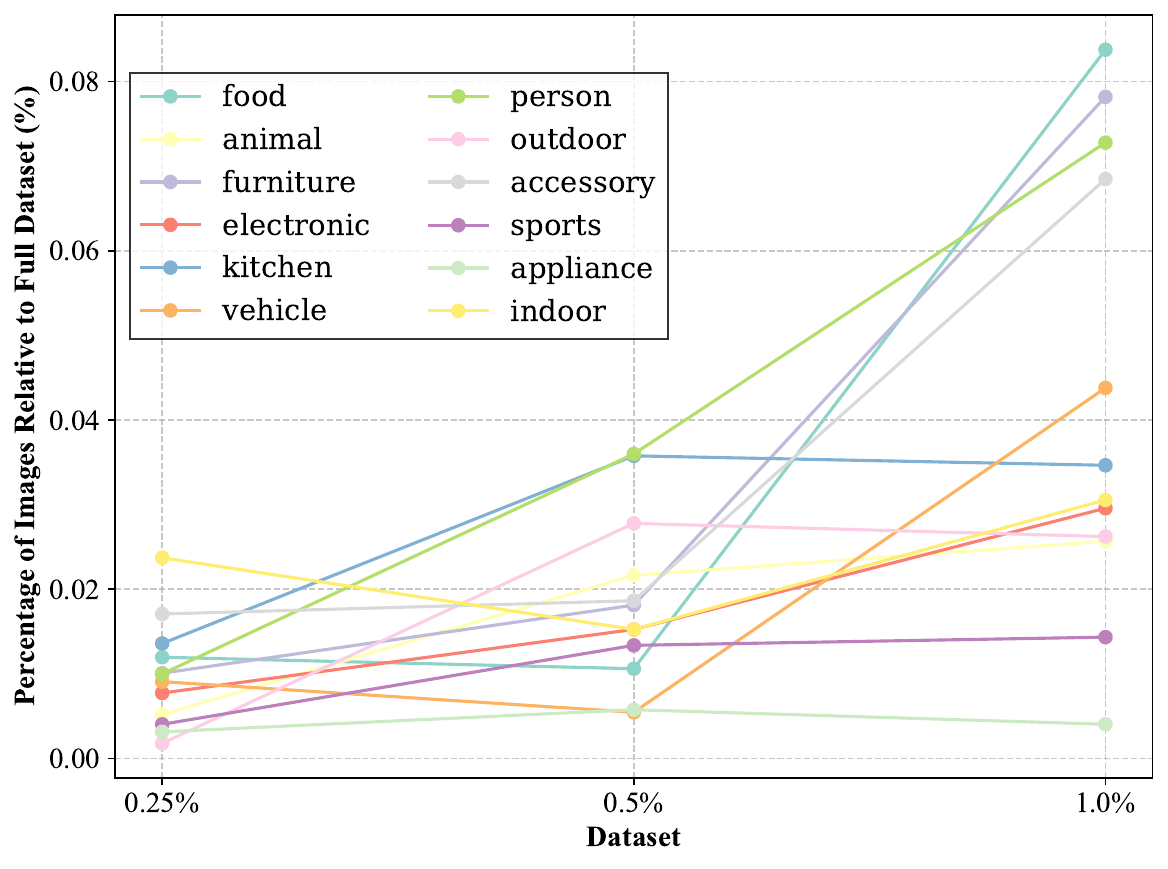}
  \end{center}
  \vspace{-0.2in}
  \caption{\textbf{Percentage of images in the distilled datasets relative to the full dataset.}}
  \label{fig:prob-distribution_2}
  \vspace{-0.2in}
\end{wrapfigure}

Table~\ref{tab:image-information} presents the distribution of images and objects across 12 supercategories in MS COCO under different \texttt{IPD} settings as well as in the original dataset. The full dataset (100\%) contains 64, 115 images and 262, 465 person-related objects, while the most compressed version (0.25\%) retains only 295 images and 5, 313 objects of this supercategory. Similar reductions are observed across all supercategories, demonstrating the significant compression effect of the \textbf{\texttt{OD\textsuperscript{3}}} distillation process. We also present the ratio of the number of images in a particular supercategory at a certain compression ratio compared to the number of images of the corresponding supercategory in the original MS COCO dataset in Fig.~\ref{fig:prob-distribution_2}.

\begin{figure}[h!]
    \centering
    \includegraphics[width=1.0\linewidth]{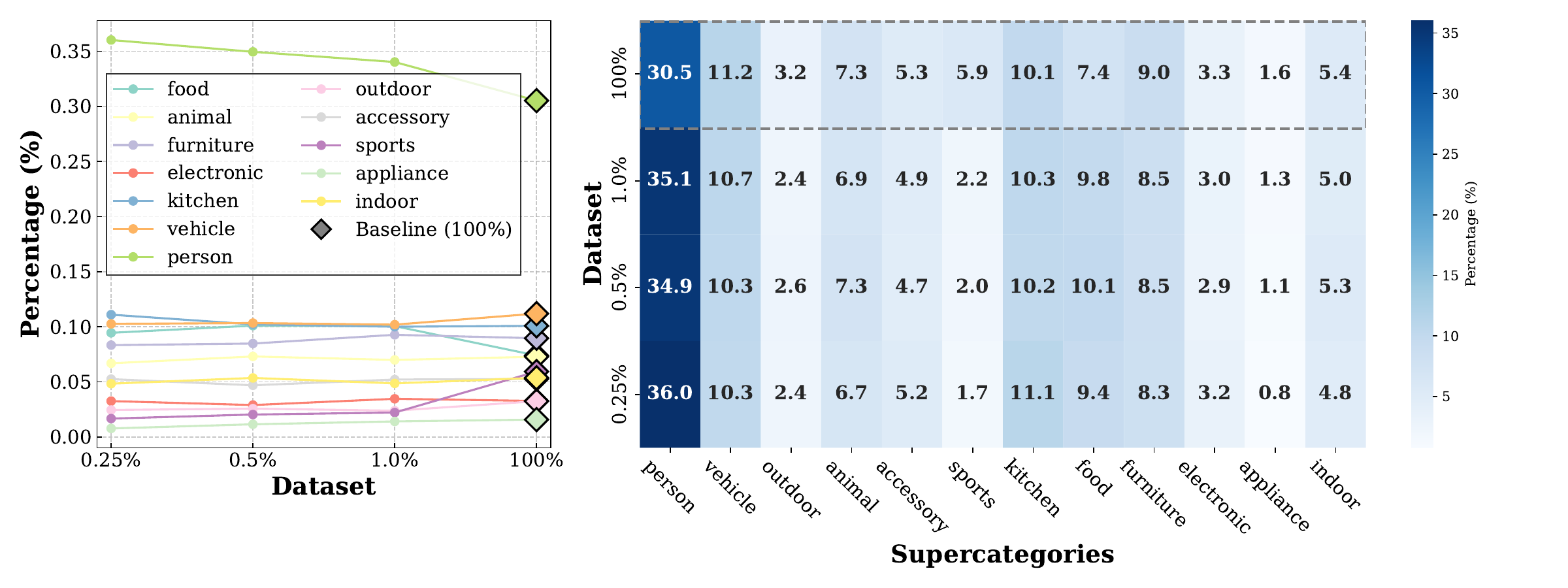}
    \caption{\textbf{Probability distribution of supercategories across datasets.} The figure highlights the relative distribution of each supercategory in the original MS COCO dataset (100\%) and its synthesized counterparts at different compression ratios (0.25\%, 0.5\%, and 1.0\%). This analysis shows that the synthesis process successfully mirrors the distribution of supercategories, $\mathcal{N}(x)$, in the original dataset. }
    \label{fig:prob-distribution}
\end{figure}

Fig.~\ref{fig:prob-distribution} further illustrates the relative probability distribution of supercategories across dataset versions. Despite significant compression that can be seen in Table~\ref{tab:image-information}, the distribution remains statistically consistent with the original dataset. This shows that \textbf{\texttt{OD\textsuperscript{3}}} does not introduce any inherent bias toward any specific category.

\begin{table}[h]
\centering
\caption{\textbf{Supercategory distribution across different \texttt{IPD} settings.} The number of images and objects per supercategory is presented for the MS COCO \citep{mscoco} dataset and the \textbf{\texttt{OD\textsuperscript{3}}} distilled versions. Note that the data in the table is the same as in Fig.~\ref{fig:prob-distribution_2}, but Fig.~\ref{fig:prob-distribution_2} displays the values as percentages. It can be seen that both the number of images and objects per supercategory are drastically compressed. Supercategory-level data is provided instead of fine-grained categories to maintain clarity and simplify comparisons.}
\resizebox{\linewidth}{!}{%
\begin{tabular}{c|c|cccccccccccc}
\toprule
\multirow{2}{*}{\textbf{IPD}} & \multirow{2}{*}{\textbf{Type}}& \multicolumn{12}{c}{\textbf{Supercategory in MS COCO}}  \\\cline{3-14}
& & \textbf{person}  & \textbf{indoor} & \textbf{food} & \textbf{kitchen} & \textbf{appliance} & \textbf{furniture} & \textbf{vehicle} & \textbf{animal} & \textbf{electronic} & \textbf{accessory} & \textbf{outdoor} & \textbf{sports} \\\midrule
\multirow{2}{*}{\makecell{100\%\\(Full Dataset)}} & Images & 64115 & 15847 & 16255 & 20792 & 7880 & 29481 & 27358 & 23989 & 12944 & 17691 & 12880 & 23218 \\
& Objects & 262465 & 46088 & 63512 & 86677 & 13479 & 76985 & 96212 & 62566 & 28029 & 45193 & 27855 & 50940 \\\midrule
\multirow{2}{*}{1.0\%} & Images & 1183 & 733 & 660 & 948 & 260 & 1012 & 694 & 596 & 615 & 882 & 464 & 423 \\
& Objects & 20158 & 2876 & 5956 & 5922 & 831 & 5489 & 6035 & 4138 & 2047 & 3083 & 1407 & 1308 \\\midrule
\multirow{2}{*}{0.5\%} & Images & 585 & 366 & 313 & 463 & 120 & 496 & 351 & 279 & 270 & 433 & 219 & 212 \\
& Objects & 10257 & 1570 & 2963 & 2996 & 335 & 2486 & 3032 & 2142 & 848 & 1374 & 752 & 595\\\midrule
\multirow{2}{*}{0.25\%} & Images & 295 & 187 & 155 & 221 & 50 & 242 & 189 & 142 & 137 & 220 & 115 & 94 \\
& Objects & 5313 & 712 & 1393 & 1636 & 113 & 1228 & 1513 & 985 & 479 & 773 & 360 & 245 \\
\bottomrule
\end{tabular}%
}

%, where the person supercategory went from $262,465$ objects in a total of $64,115$ images, to only $5313$ objects in $295$ images.
\label{tab:image-information}
\end{table}

\section{More Ablations}

 Table~\ref{tab:bbox_extension} further illustrates the impact of SA-DCE on object detection performance. Our SA-DCE method consistently outperforms both our no-extension and static extension baseline methods. Notably, it improves mAP scores while striking a balance between small and medium object detection. The no-extension setting suffers from reduced performance on small objects due to limited contextual information, whereas static extension provides slight improvements but lacks adaptability to object scale. In contrast, SA-DCE dynamically adjusts the context extension based on object size, leading to significant gains, particularly in small-object detection. These results demonstrate that SA-DCE effectively enhances detection robustness while preserving overall performance across different object scales.

\begin{table}[h!]
\centering
\caption{\textbf{Ablation Study of SA-DCE.} The table evaluates the influence of extending statically and dynamically (using SA-DCE) the bounding boxes (in pixels) of the objects in the distilled dataset across varying compression ratios on the MS COCO performance. Static extension refers to applying constant extension $\bm \bar{r}$ to all inserted objects regardless of their size. We set $\bm \bar{r}$ as 10 pixels.}
\resizebox{0.75\linewidth}{!}{%
\begin{tabular}{c|c|cccccc}
\toprule
\textbf{\texttt{IPD}} &  \makecell{\textbf{Extension} \\ \textbf{(pixels)}} & \textbf{mAP} & \textbf{mAP$_{50}$} & \textbf{mAP$_{75}$} & \textbf{mAP$_s$} & \textbf{mAP$_m$} & \textbf{mAP$_l$} \\ 
\midrule
\multirow{3}{*}{\textbf{0.5\%}} 
& No extension  & 16.60 & 30.30 & 16.30 &6.80 &21.70  &\textbf{23.30}\\
& Static extension  & 16.80 & 30.70 & 16.60 &7.40  &22.10 &22.90      \\
& SA-DCE & \textbf{17.20} & \textbf{31.90} & \textbf{16.90} & \textbf{8.40} & \textbf{23.00} & 22.80 \\\midrule
\multirow{3}{*}{\textbf{1.0\%}} 
& No extension & 22.00 & 38.70 & 22.30 &9.70  &27.40  &30.10   \\
& Static extension & 22.30 & 38.80 & 22.90 &9.90&27.40 &\textbf{30.40}   \\
& SA-DCE & \textbf{22.40} & \textbf{39.50} & \textbf{22.90} & \textbf{10.60} & \textbf{28.00} & 29.80\\ \bottomrule
\end{tabular}
}
\label{tab:bbox_extension}
\end{table}

Table~\ref{tab:conf-threshold} highlights the effect of varying the confidence threshold ($\eta$) on detection performance. Setting $\eta = 0.2$ consistently yields the best overall results across different \texttt{IPD} values, improving mAP and balancing small, medium, and large object detection. Lower thresholds ($\eta = 0.1$) allow more candidates but introduce noise, while higher thresholds ($\eta \geq 0.3$) remove potentially useful detections, leading to a drop in performance. These findings demonstrate that careful tuning of $\eta$ is crucial for optimizing detection accuracy.

\begin{table}[h!]
\centering
\caption{\textbf{Ablation Study of Confidence Threshold ($\eta$).} Objects with confidence lower than $\eta$ (determined by observer model) are removed in the candidate screening stage.}
\resizebox{0.76\linewidth}{!}{%
\begin{tabular}{c|c|cccccc}
\toprule
\textbf{\texttt{IPD}} &  \textbf{\makecell{\textbf{Confidence } \\ \textbf{Threshold ($\eta$)}}} & \textbf{mAP} & \textbf{mAP$_{50}$} & \textbf{mAP$_{75}$} & \textbf{mAP$_s$} & \textbf{mAP$_m$} & \textbf{mAP$_l$} \\ 
\midrule
\multirow{5}{*}{\textbf{0.25\%}} 
& 0.1  &11.10 & 21.70 & 10.50 & 5.30 & 15.50 &14.50   \\
& 0.2  & \textbf{12.90} &\textbf{24.30} & \textbf{12.10} &\textbf{5.60} & \textbf{16.80} & \textbf{17.70}  \\
& 0.3 & 10.50 & 21.00 & 9.40& 4.90 & 14.80 & 13.40    \\
& 0.4 & 10.00 & 19.80 & 9.00& 5.40 & 13.80 & 12.90 \\
& 0.5 & 10.20 & 20.30 &  9.30 & 5.00 & 14.50 & 12.90 \\ 
\midrule

\multirow{5}{*}{\textbf{0.5\%}} 
& 0.1  &17.00 & 31.50 & 16.60 & 8.10 & 22.80 & 22.60 \\
& 0.2  &  \textbf{17.20} & \textbf{31.90} & \textbf{16.90} & \textbf{8.40} & \textbf{32.00} & \textbf{22.80} \\
& 0.3 &  16.20 & 30.10 & 15.90 & 7.30 & 21.60 & 21.60 \\
& 0.4 & 16.40 & 30.60 & 16.10 & \textbf{8.40} & 22.50 & 21.90   \\
& 0.5 & 15.70 & 29.50 & 15.20 & 7.20 & 21.30 & 20.50   \\\midrule

\multirow{5}{*}{\textbf{1.0\%}} 
& 0.1  & 21.70 & 38.30 & 22.20 &\textbf{10.80} &27.80& 28.90  \\
& 0.2  &  \textbf{22.40} & \textbf{39.50} & \textbf{22.90} & 10.60 & \textbf{28.00} & \textbf{29.80}  \\
& 0.3 & 22.00 & 39.00 & 22.30 & 10.60 & 27.60 & 29.30 \\
& 0.4 & 22.00 & 38.80 & 22.40 & 10.20 & 27.80 & 28.90\\ 
& 0.5 & 21.80 & 38.50 & 22.20 & 10.30 & 27.70 & 29.00 \\ 
\bottomrule
\end{tabular}
}
\label{tab:conf-threshold}
\end{table}

Table~\ref{tab:canvas_size} analyzes the effect of canvas size on detection performance across different \texttt{IPD} values. The canvas size was selected based on the average width and height of all training images in the MS COCO dataset, with additional smaller and larger canvas sizes included for comparison and evaluation. Results indicate that while an optimal canvas size (\(484\times578\)) achieves the highest mAP scores, further reduction in canvas dimensions leads to a drop in performance. This suggests that excessively small canvases limit the available contextual information, negatively impacting detection accuracy. Conversely, overly large canvases introduce unnecessary background noise, reducing effectiveness. These findings highlight the importance of selecting a balanced canvas size to maximize object representation while maintaining relevant contextual cues for dataset distillation.
\begin{table}[h!]
\centering
\caption{\textbf{Ablation Study of Canvas Size.} The table evaluates the influence of canvas size on the MS COCO performance of the distilled dataset across varying compression ratios.}
\resizebox{0.76\linewidth}{!}{%
\begin{tabular}{c|c|cccccc}
\toprule
\textbf{\texttt{IPD}} &  \makecell{\textbf{Canvas} \\ \textbf{Size (pixels)}} & \textbf{mAP} & \textbf{mAP$_{50}$} & \textbf{mAP$_{75}$} & \textbf{mAP$_s$} & \textbf{mAP$_m$} & \textbf{mAP$_l$} \\ 
\midrule
\multirow{4}{*}{\textbf{0.25\%}} 
&$363\times433$  &10.30 &20.80&9.10&4.80&13.60&14.20\\
& $484\times578$ &\textbf{12.90}&\textbf{24.30}&\textbf{12.10}&\textbf{5.60}&\textbf{16.80}  &\textbf{17.70}\\
& $726\times867$  & 10.50  &20.50&9.50 &4.70  &16.30  &12.50\\
&$968\times1156$  &8.80 &17.50&7.80&3.60&14.90&10.20\\ \midrule
\multirow{4}{*}{\textbf{0.5\%}} 
&$363\times433$&15.40 & 29.10 & 14.70 & 7.20 & 19.90 & 21.50  \\
&$484\times578$  & \textbf{17.20} & \textbf{31.90} & \textbf{16.90} & \textbf{8.40} & \textbf{23.00} & \textbf{22.80}\\
&$726\times867$ &15.70&29.10&15.20&7.30&22.60&19.90\\
&$968\times1156$&13.70&26.00&12.90&7.00&20.00&16.60\\ \midrule
\multirow{4}{*}{\textbf{1.0\%}} 
&$363\times433$& 20.90 & 37.20 & 21.10 & 9.80 & 25.70 & 28.30 \\

&$484\times578$  & \textbf{22.40} & \textbf{39.50} & \textbf{22.90} & 10.60 & \textbf{28.00} & \textbf{29.80}\\
&$726\times867$  &21.00&37.30&21.40 &\textbf{10.80} &27.60&26.40  \\
&$968\times1156$   &16.80&30.40&16.90&8.40&24.10&20.40\\
\bottomrule
\end{tabular}
}
\label{tab:canvas_size}
\end{table}

\begin{figure}
    \centering
    \includegraphics[width=0.6\linewidth]{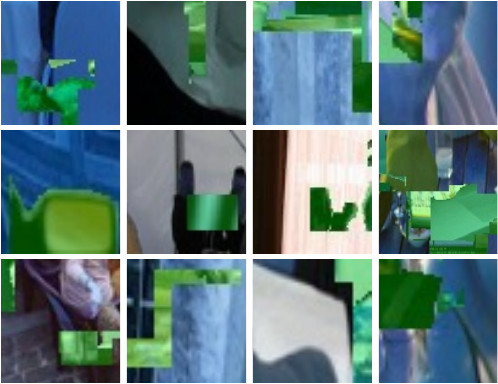}
    \caption{\textbf{Qualitative results of the candidate screening process of \textbf{\texttt{OD\textsuperscript{3}}} on MS COCO.} The screened and removed objects are highlighted with a light green mask. It can be seen that these low-confidence objects are unreliable and can be regarded as ``visual noise''.}
    \label{fig:removed-obj}
\end{figure}

Figure \ref{fig:removed-obj} illustrates samples of objects removed by the candidate-screening module during the distillation process. Objects with low confidence are identified and masked (shown in light green), revealing how these regions often correspond to ambiguous shapes, partial objects, or background clutter. Removing these ``visual noise'' segments helps the distilled dataset focus on high-quality, semantically stable content, which in turn strengthens the downstream learning signals.

\begin{table}[t!]
\caption{\textbf{More Cross-Architecture Results with a DETR-based target model (0.5\% IPD).} These experiments highlight a trend where transformer-based detectors are substantially more data-hungry than their CNN counterparts.}
\centering
\resizebox{0.7\linewidth}{!}{%
{
\begin{tabular}{c|c|c|c|c}
\toprule
\textbf{Observer} & \textbf{Target} & \textbf{mAP (\%)} & \textbf{mAP$_{50}$ (\%)} & \textbf{mAP$_{75}$ (\%)} \\
\midrule
Faster R-CNN        & Deformable DETR   & 6.00  & 12.20 & 5.50 \\
Deformable DETR     & Deformable DETR   & 9.40  & 18.80 & 8.60 \\
\midrule
Faster R-CNN        & Faster R-CNN       & 17.20 & 39.50 & 22.90 \\
\bottomrule
\end{tabular}%
}}
\label{tab:cross-architecture-transformers}
\end{table}

We extend the cross-architecture evaluation from Table \ref{tab:cross_architecture} in Table \ref{tab:cross-architecture-transformers}. Results show that DETR-based models degrade sharply under extreme data scarcity, regardless of the nature of the observer model used, whereas Faster R-CNN maintains substantially stronger performance. DETR and its variants are known to require large datasets, long optimization schedules, and abundant object diversity to converge well \citep{deittowards}. When the distilled dataset is extremely small, these models struggle to learn effective queries and attention patterns. This may represent a valuable direction for future exploration in the DD field.

\section{Proof of Theorem~\ref{the:main}}
\label{sec:proof_thoerem}
\begin{proof}
Let $t\in\mathbb{N}$ be the current iteration index with $0 \le t < T$. We assume for this iteration that objects placed on the canvas $\mathbf{x}_t$ do not overlap. Let the canvas $\mathbf{x}_t$ contain $K$ objects $\{o_r\}_{r=0}^K$. We sort these objects according to their confidence scores $q(o_r)$ and partition them into two sets based on a threshold $\eta$:
\begin{equation}
\begin{aligned}
    \bigl\{o_{i_0}, o_{i_1}, \dots, o_{i_M}\bigr\}
\quad\text{where}\quad
q(o_{i_0}) \;\le\;q(o_{i_1})\;\le \dots \le\; q(o_{i_M}) \;<\;\eta, \\
\bigl\{o_{i_{M+1}}, \, o_{i_{M+2}}, \dots, \, o_{i_K}\bigr\}
\quad\text{where}\quad
\eta \;\le\;q(o_{i_{M+1}})\;\le \dots \le\;q(o_{i_K}).
\end{aligned}
\end{equation}
The first set satisfy $\mathbb{E}_r\left[p\left(q\left(o_{i_r}\right)<\eta\right)\right]=\frac{M+1}{K}$. Applying the removal operator $f_{\mathrm{remove}}(\mathbf{x}_t)$ discards every object whose confidence is below $\eta$, i.e., $\bigl\{o_{i_0}, o_{i_1}, \dots, o_{i_M}\bigr\}$.

Hence, the updated canvas $\mathbf{x}_{t+1}$ preserves only those objects whose scores exceed $\eta$, and it may then be ``refilled'' by $f{\mathrm{add}}(\cdot)$ with new (randomly synthesized) objects from the same distribution as in previous iterations.

Then, we can compare the $\Phi(\mathbf x)$ of $\mathbf x_t$ and $\mathbf x_{t+1} . \Phi(\mathbf x)$ can be described as
\begin{equation}
    \Phi(\mathbf x)=\frac{\sum_{j=0}^M s\left(o_{i_j}\right) q\left(o_{i_j}\right)+\sum_{j=M+1}^K s\left(o_{i_j}\right) q\left(o_{i_j}\right)}{\sum_{j=0}^M s\left(o_{i j}\right)+\sum_{j=M+1}^K s\left(o_{i_j}\right)}
\end{equation}
where $\sum_{j=M+1}^K s\left(o_{i_j}\right)$ and $\sum_{j=M+1}^K s\left(o_{i_j}\right) q\left(o_{i_j}\right)$ are same for $\mathbf x_t$ and $\mathbf x_{t+1}$. In general, we will
fill the canvas at each iteration, so $\sum_{j=0}^M s\left(o_{i_j}\right)$ can also be considered constant. And the difference between $\mathbf x_t$ and $\mathbf x_{i+1}$ is $\frac{\sum_{j=0}^M s\left(o_{i_j}\right) q\left(o_{i_j}\right)}{S}$, where $S$ is the areas of the canvas. Due to $\mathbb{E}_{0\leq j \leq M}\left[p\left(q\left(o_{i_j}\right)<\eta\right)\right]=1$ for $\mathbf x_t$, we can get $p\left(\mathbb{E}_{0 \leq j \leq M}\left[q\left(o_{i_j}\right)\right]-\mathbb{E}[\eta] \geq 0\right)=0$, and $\mathbb{E}_{0\leq j \leq M}\left[p\left(q\left(o_{i_j}\right)<\eta\right)\right]=\frac{M+1}{K}$ for $\mathbf x_{t+1}$. Then, we can get
\begin{equation}
    p\left(\mathbb{E}_{0 \leq j \leq M}\left[q\left(o_{i_j}\right)\right]-\mathbb{E}[\eta] \geq 0\right)=\frac{K-M-1}{K}
\end{equation}
Since object is uniformly distributed, so $p$ and $\mathbb{E}$ are able to swap places.
Because $p\left(\mathbb{E}_{0 \leq j \leq M}\left[q\left(o_{i_j}\right)\right] \geq \eta\right)=\frac{K-M-1}{K}$ for $\mathbf x_{i+1}$, we can prove that
\begin{equation}
   \Phi\left(\mathbf x_T\right) \geq \Phi\left(\mathbf x_{T-1}\right) \geq \Phi\left(\mathbf x_{T-2}\right) \geq \cdots \geq \Phi\left(\mathbf x_0\right) 
\end{equation}
{\bf Consistency of $\mathcal{N}(\mathbf x)$ and overlaps.}
As $T \rightarrow \infty$, the canvas becomes fully populated in both the {\em add-only} and {\em add-then-remove} strategies, so the number of objects $\mathcal{N}\left(\mathbf x_T\right)$ is generally similar (i.e., both can fill the canvas to full capacity).

{\em Overlap handling.} 

When $T$ is sufficiently large, the canvas will necessarily be filled, so it can be assumed that the first form and the second form of $\mathcal{N}\left(\mathbf x_T\right)$ are consistent. So $G_2$ remains greater than $G_1$.  When there are some overlaps of objects in the iteration, the conclusion still holds. For example, object $o_a$ and $o_b$ overlap, and their overlap region is $o_d$. The score of $o_d$ is between $q\left(o_a\right)$ and $q\left(o_b\right)$. If both $q\left(o_a\right)$ and $q\left(o_b\right)$ are larger or smaller than $\eta$, then both of them will not be considered. If one is larger and one smaller than $\eta$ (assuming that $q\left(o_a\right) \leq \eta$ and $\left.q\left(o_b\right) \geq \eta\right)$, then $o_a$ is removed and $o_c$ will also be removed in the process of {\em screening}, and the portion left behind (i.e., possibly the mutilated $o_b \rightarrow \hat{o}_b$ ) may not be detectable by the detector, or it may be successfully detected. Even assuming that this is undetectable for $\hat{o}_b$ (i.e., the confidence score is low), then in the next iteration it will still be removed.

Assume that the probability of having no overlap with another object is $p_1$. The probability that $q\left(\hat{o}_b\right) \leq \tau$ is detected will be $p_2$. This probability of it being removed or not having an overlap in the next iteration is $p_1+\left(1-p_1\right) p_2$, which is consistently greater than $\left(1-p_1\right)\left(1-p_2\right)$ when $p_2 \geq 0.5$. If $p_2<0.5$, this means that $\hat{o}_b$ is a qualified sample (detectable by detector or observer) and therefore does not need to be removed.

Thus, in the presence of overlap, $G_2$ remains greater than $G_1$.

\end{proof}
%%%%%%%%%%%%%%%%%%%%%%%%%%%%%%%%%%%%%%%%%%%%%%%%%%%%%%%%%%%%

\section{Use of Large Language Models}

An LLM was used only to refine the writing of the paper. All ideas, methods, experimental designs, and results were conducted by the authors.

\end{document}